\documentclass{article}

\usepackage{microtype}
\usepackage{graphicx}
\usepackage{subcaption}
\usepackage{booktabs} 

\usepackage{hyperref}

\usepackage[accepted]{icml2025}

\usepackage{amsmath}
\usepackage{amssymb}
\usepackage{mathtools}
\usepackage{amsthm}
\usepackage{listings}
\usepackage{url}

\usepackage{bbm}
\usepackage{colortbl}
\usepackage{placeins} 

\newcommand{\E}{\mathbb{E}}

\DeclareMathOperator*{\argmax}{arg\,max}
\DeclareMathOperator*{\argmin}{arg\,min}

\DeclareMathOperator{\action}{a}
\DeclareMathOperator{\state}{s}
\DeclareMathOperator{\reward}{r}

\definecolor{ellipsescolor}{gray}{0.5} 
\definecolor{highlightcolor}{rgb}{1.0,1.0,0.6}
\definecolor{highlightcolornn}{rgb}{1.0,1.0,0.6}
\definecolor{highlightgcolorgbrl}{rgb}{0.6, 1.0, 0.6}
\definecolor{skipcodecolor}{gray}{0.7} 

\usepackage[capitalize,noabbrev]{cleveref}

\theoremstyle{plain}

\theoremstyle{definition}

\theoremstyle{remark}

\usepackage[textsize=tiny]{todonotes}

\definecolor{cb-rose}       {RGB}{255, 109, 182}

\definecolor{tabgreen}{RGB}{76, 175, 80}
\definecolor{tabred}{RGB}{227, 26, 28}

\icmltitlerunning{Gradient Boosting Reinforcement Learning}

\begin{document}

\twocolumn[
\icmltitle{Gradient Boosting Reinforcement Learning}



\icmlsetsymbol{equal}{*}

\begin{icmlauthorlist}
\icmlauthor{Benjamin Fuhrer}{1}
\icmlauthor{Chen Tessler}{2}
\icmlauthor{Gal Dalal}{2}
\end{icmlauthorlist}

\icmlaffiliation{1}{NVIDIA, Tel-Aviv, Israel}
\icmlaffiliation{2}{NVIDIA Research, Tel-Aviv, Israel}

\icmlcorrespondingauthor{Benjamin Fuhrer}{bfuhrer@nvidia.com}
\icmlcorrespondingauthor{Chen Tessler}{ctessler@nvidia.com}
\icmlcorrespondingauthor{Gal Dalal}{gdalal@nvidia.com}

\icmlkeywords{Machine Learning, Reinforcement Learning, Gradient Boosting Trees, Gradient Boosting, ICML}

\vskip 0.3in
]



\printAffiliationsAndNotice{}  

\begin{abstract}

We present Gradient Boosting Reinforcement Learning (GBRL), a framework that adapts the strengths of gradient boosting trees (GBT) to reinforcement learning (RL) tasks. While neural networks (NNs) have become the de facto choice for RL, they face significant challenges with structured and categorical features and tend to generalize poorly to out-of-distribution samples. These are challenges for which GBTs have traditionally excelled in supervised learning. However, GBT's application in RL has been limited. The design of traditional GBT libraries is optimized for static datasets with fixed labels, making them incompatible with RL's dynamic nature, where both state distributions and reward signals evolve during training. GBRL overcomes this limitation by continuously interleaving tree construction with environment interaction. Through extensive experiments, we demonstrate that GBRL outperforms NNs in domains with structured observations and categorical features while maintaining competitive performance on standard continuous control benchmarks. Like its supervised learning counterpart, GBRL demonstrates superior robustness to out-of-distribution samples and better handles irregular state-action relationships.
\end{abstract}

\section{Introduction}
\label{sec:introduction}

Many real-world decision-making tasks involve structured observations, where data can be organized in a tabular format and follow predefined organizational patterns. Unlike unstructured data (such as images or raw sensor data), structured observations often include heterogeneous features, both numerical and categorical. These features carry direct semantic meaning that can be used for prediction without complex feature extraction. Such structured observations closely resemble tabular datasets commonly encountered in supervised learning. Notable examples where structured observations are crucial include recommendation systems, healthcare diagnostics, digital advertising, fraud detection, weather prediction, and financial trading.


In these domains, a key challenge lies in handling structured observations, where the importance of different components varies dynamically with the agent's state and task. Neural network (NN)-based solutions struggle with such data, requiring sophisticated architectures and preprocessing techniques to capture structural relationships \cite{zabergja2024tabulardataattentionneed, kadra2021welltunedsimplenetsexcel, arik2020tabnetattentiveinterpretabletabular, gorishniy2023revisiting, hollmann2023tabpfntransformersolvessmall}. This often results in complex models that sacrifice sample efficiency and generalization capability.

Gradient boosting trees (GBT) offer a promising alternative through their natural handling of structured data. Their success in supervised learning stems from an iterative ensemble-building process, in which each decision tree refines predictions by partitioning the input space along individual features. This approach has made GBT frameworks such as XGBoost \cite{xgboost}, LightGBM \cite{lightgbm}, and CatBoost \cite{catboost} integral in domains such as finance \cite{TIAN2020150}, healthcare \cite{gbt_federated_healthcare, gbt_healthcare, gbt_reliability_healthcare}, and competitive data science \cite{xgboost_demo}.



This approach creates an ensemble of piecewise constant functions that excel at capturing non-smooth patterns and abrupt transitions in data \cite{jeffares2024deeplearningtelescopinglens, Beyazit2023AnIB, grinsztajn2022treebased}. These properties are particularly valuable in RL, where state-action relationships often exhibit sudden changes, for instance, when transitioning between different phases of a task or when certain state variables cross critical thresholds. Such discontinuous relationships are common in structured RL tasks, from inventory management where actions change discretely with stock levels, to game scenarios where strategic decisions shift abruptly based on state conditions.


Despite these potential benefits, adapting GBT to RL presents significant challenges. Traditional GBT frameworks are designed for static datasets with stable feature distributions and predefined loss functions, where the focus is on optimizing batch training efficiency. This contrasts sharply with RL's dynamic nature, where state distributions evolve during training, data is collected iteratively, and rewards are often delayed. While traditional GBT libraries can technically be adapted to RL using custom loss functions and incremental learning, our experiments with CatBoost and XGBoost (\cref{subsec:comparison_traditional}) show that this approach fails to scale due to their supervised learning-oriented design.

We address these challenges through Gradient Boosting Reinforcement Learning (GBRL), a framework that bridges the gap between GBT's strengths and RL's unique requirements. Our contributions are:

\textbf{1. New framework for GBT in RL:} We establish GBT as a viable function approximator for RL by devising popular algorithms---PPO, A2C, and AWR---with GBT backends. In popular environments, our framework demonstrates competitive performance against NNs while showing particular strengths on structured tasks.

\textbf{2. Improved robustness:} We demonstrate that GBT's inherent strengths translate effectively to RL tasks. Through extensive dedicated experiments, we show superior robustness to out-of-distribution scenarios, resilience to spurious correlations, and better handling of state-space perturbations -- critical advantages over NNs for real-world applications.

\textbf{3. Practical implementation:} We provide a CUDA-accelerated \cite{nvidia_cuda} implementation that seamlessly integrates with existing RL libraries like Stable-baselines3 \cite{stable-baselines3}\footnote{The GBRL core library is available at \url{https://github.com/NVlabs/gbrl}.}
\footnote{Actor-Critic implementations integrated within Stable-baselines3, are available at \url{https://github.com/NVlabs/gbrl_sb3}.} Our shared Actor-Critic design achieves state-of-the-art performance while significantly reducing both memory and computational requirements on modern hardware.

\section{Related Work}

\paragraph{GBT Beyond Supervised Learning.} Recent advances have extended the capabilities of GBT beyond traditional regression and classification. In ranking problems, GBT has been used to directly optimize ranking metrics \cite{lyzhin2023tricks}, as demonstrated by frameworks such as StochasticRank \cite{ustimenko2020stochasticrank}. Furthermore, GBT offers probabilistic predictions through frameworks such as NGBoost \cite{duan2020ngboost}, allowing uncertainty quantification \cite{malinin2021uncertainty}. The connection between GBT and Gaussian processes \cite{catboost_gp, sigrist2022gaussian} offers further possibilities for uncertainty-aware modeling. Recently, \citet{ivanov2021boost} modeled graph-structured data by combining GBT with graph NNs.

Despite their versatility, applying GBT in RL remains a relatively less explored area. Several works have employed GBT as a function approximator within off-policy RL methods, including its use in Q-learning \cite{abel2016exploratory} and in bandit settings to learn inverse propensity scores \cite{pmlr-v206-london23a}. Recently, \citet{brukhim2023boosting} proposed a boosting framework for RL where a base policy class is incrementally enhanced using linear combinations and nonlinear transformations, forming a 2-layer NN. However, these previous works have not demonstrated scalability and effectiveness in complex, high-dimensional RL environments that require extensive interactions.
In this work, we show how to adapt the framework of GBT to successfully solve large-scale RL problems for the first time.

\paragraph{Tabular Data.} Previous work in RL has focused predominantly on the use of NNs due to their ability to capture complex patterns in high-dimensional data. Techniques such as Q-learning and Actor-Critic methods have advanced significantly, demonstrating success in tasks involving raw sensory inputs like images, text, and audio. However, NNs that perform well on tabular data typically have very specialized architectures \cite{katzir2021netdnf, somepalli2021saint, gorishniy2023revisiting, arik2020tabnetattentiveinterpretabletabular} and are different from the standard multi-layer perceptrons that are often used in RL tasks and algorithms \cite{ota2021training}. Even with specialized architectures, GBT often performs equally or better on tabular datasets \cite{Borisov_2024, NEURIPS2023_f06d5ebd, grinsztajn2022treebased, shwartzziv2021tabulardatadeeplearning}. GBT's success can be attributed to a key difference in inductive biases compared to NNs. \citet{jeffares2024deeplearningtelescopinglens} have shown that the kernel representations induced by GBTs are bounded and often behave more predictably on irregular or out-of-distribution inputs. In contrast, NN-tangent kernels can be unbounded and vary significantly on test points far from the training distribution.

\paragraph{Policy Optimization through Functional Gradient Ascent.} \citet{nonparametricrl} introduced Non-Parametric Policy Gradients (NPPG). By combining the policy gradient theorem \cite{NIPS1999_464d828b} with functional gradient ascent, NPPG directly optimizes the policy. Although NPPG sets the foundation for combining GBT with RL, it is limited to discrete action spaces and policies represented by the Gibbs distribution.
GBRL extends the ideas laid in NPPG to general Actor-Critic algorithms and generalizes to any policy representation and objective. 

\begin{figure*}[t!]
\centering
    \includegraphics[width=\linewidth]{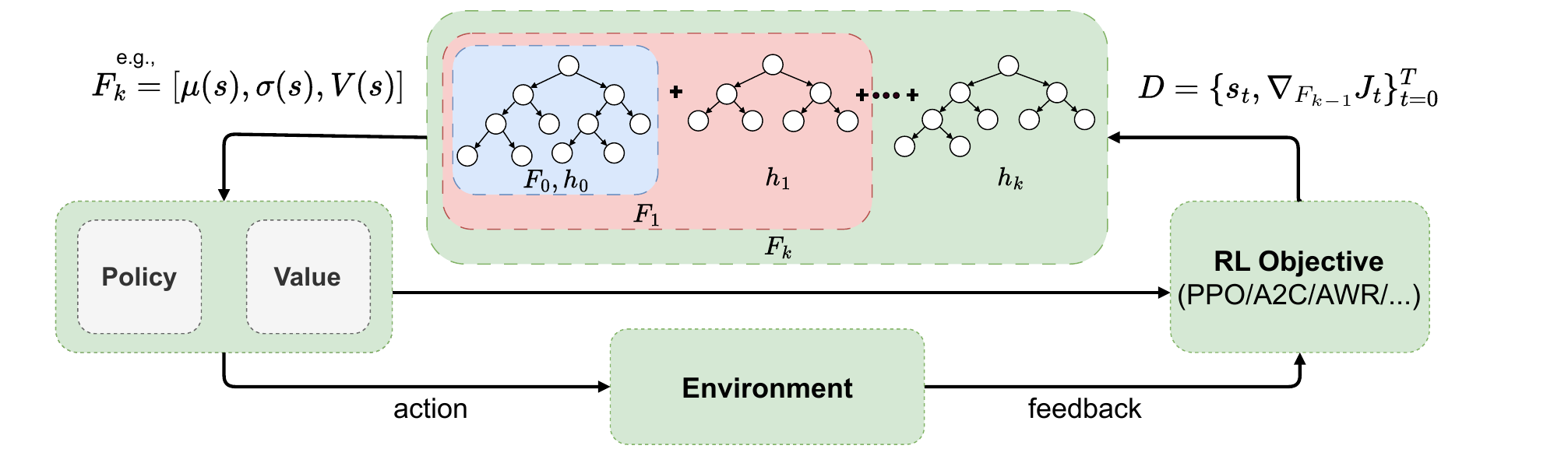}
    \caption{\textbf{The GBRL framework.} 
    The actor's policy and critic's value function are parameterized by the tree ensemble \(F_k\). For example, \(F_k(\state) = [\mu(\state), \sigma(\state), V(\state)]\) for a Gaussian policy. Starting from \(F_0\), at each training iteration, \(k\), GBRL collects a rollout and computes the gradient \(\nabla_{F_{k-1}}J(\pi_{F_{k-1}})\) with respect to the current ensemble. This gradient is then used to fit the next tree. Adding the tree to the ensemble updates it to \(F_k(\state) = F_{k-1}(\state) + \epsilon h_k(\state)\), where \(\epsilon\) is the learning rate.}
    \label{fig:gbrl_diagram}
\end{figure*}
 
\section{RL Preliminaries}
A fully observable infinite-horizon MDP is characterized by the tuple \( (\mathcal{S}, \mathcal{A}, P, \mathcal{R})\). At each step, the agent observes a state \( \state \in \mathcal{S} \) and samples an action \(\action \in \mathcal{A}\) from its policy \(\pi(\state, \action)\). The action causes the system to transition to a new state \( \state'\) based on the transition probabilities \(P(\state' | \state, \action)\), and produce a reward \( \reward \sim \mathcal{R} (\state, \action) \). The agent's objective is to 
maximize the expected discounted reward \(J(\pi) = \E[\sum_{t=0}^\infty \gamma^t \reward_t]\), with a discount factor \( \gamma \in [0, 1)\).

The state-action value function \(Q_\pi(\state, \action) := \E_\pi[\sum_{t=0}^\infty \gamma^{t}\reward_t | \state_0 = \state, \action_0 = \action]\) estimates the expected returns of performing action \(\action\) in state \( \state \) and then acting according to \(\pi\). Additionally, the value function \(V_\pi(\state) := \E_\pi[\sum_{t=0}^\infty \gamma^{t}\reward_t | \state_0 = \state]\), predicts the expected return starting from state \(\state\) and acting according to \(\pi\). Finally, the advantage function \(A_\pi(\state, \action) := Q_\pi(\state, \action) - V_\pi(\state)\), indicates the expected relative benefit of performing action \(\action\) over acting according to \(\pi\).

Actor-Critic methods are a common approach in RL. They simultaneously learn a policy and a value function to improve the efficiency and stability of training. For conciseness, in the body of the paper we focus on Proximal Policy Optimization (PPO) \cite{ppo} which jointly trains an actor and critic. In the supplementary material, we also showcase the implementation and results for A2C \cite{a2c} and AWR \cite{awr}.


\section{Gradient Boosting Reinforcement Learning}
Deep RL with NNs faces significant challenges when dealing with structured observations containing both numerical and categorical features. Similarly to tabular datasets in supervised learning, these observations consist of heterogeneous features representing fundamentally different types of information, such as numerical measurements alongside categorical attributes describing system states or task conditions. These features create highly irregular patterns, as they may independently affect the environment based on their context, scale, or statistical properties. NNs struggle to model such patterns effectively \cite{Beyazit2023AnIB, jeffares2024deeplearningtelescopinglens}, and their performance heavily depends on preprocessing methods \cite{Hancock2020, gorishniy2023embeddingsnumericalfeaturestabular}. In contrast, GBT naturally handles irregular patterns and structured feature spaces efficiently. 
To leverage this strength, we present GBRL, an adaptation of GBT to RL tasks with such representations that frequently arise in real-world problems.

\subsection{GBT as Functional Gradient Descent}
GBT \cite{10.1214/aos/1013203451} combines decision tree ensembles with functional gradient descent to learn complex nonlinear functions \cite{functional_gd}. Given a loss function
\(L\) and a dataset, \(D = \{(\mathbf{x}_i, \mathbf{y}_i)\}_{i=1}^N\), where \(\mathbf{x}_i \in \mathbb{R}^M\) and \(\mathbf{y}_i \in \mathbb{R}^D \), GBT aims to learn a function \(F\), mapping inputs \(\mathbf{x}\) to outputs \(\mathbf{y}\). Through functional gradient descent, GBT minimizes the expected loss \(\mathbb{E}_{\mathbf{x}, \mathbf{y}}[L(\mathbf{y}, F(\mathbf{x}))]\) with respect to the learned function \(F\). Unlike parametric gradient descent, which updates parameters, functional gradient descent updates a learned function directly.

Starting with an initial function \(F_0\). At each iteration \(k\), GBT refines \(F_{k-1}\), by taking a step in the negative direction of the gradient \(g_k := \nabla_{F_{k-1}}L(\mathbf{y}, F_{k-1}(\mathbf{x}))\). The negative gradient is projected onto the space of possible trees \(h \in \mathcal{H}\) by constructing a new binary decision tree that minimizes the following objective:
\begin{equation} 
\label{eq: h min}
h_k = \arg\min_{h \in \mathcal{H}} \| -g_k  - h(\mathbf{x}) \|_2^2 \, .
\end{equation}
  
Through this process, GBT minimizes the expected loss, resulting in the additive model:
\begin{equation}
F_K(\mathbf{x}) = F_0 + \sum_{k=1}^{K}\epsilon h_k(\mathbf{x}) \, ,
\end{equation}
where, \(\epsilon\) is the learning rate controlling the step size in each iteration.

While standard GBT frameworks excel at handling structured data in supervised tasks, they require a tailored solution for RL’s iterative, dynamic, and multi-objective nature.

\subsection{The GBRL Framework}

\begin{table}[]
\centering
\resizebox{\linewidth}{!}{%
\begin{tabular}{lcc}
\toprule
\textbf{Library}  & \textbf{Incremental Learning (sec)} & \textbf{Standard Learning (sec)} \\ \midrule
CatBoost & $36.17 \pm 2.77$ & $1.15 \pm 0.03$ \\
XGBoost  & $67.11 \pm 0.70$ & $2.26 \pm 3.70$ \\ 
GBRL     & $\mathbf{5.47} \pm \mathbf{1.19}$ & N/A \\
\bottomrule
\end{tabular}%
}
\caption{\textbf{Incremental Learning Speed in GBT Libraries.} Training times (seconds) for 1000 boosting iterations on random batches of 128 samples with 20 input features and 3 targets. While CatBoost and XGBoost train efficiently in standard mode using the same batch for all iterations, their training time increases significantly in incremental learning, where each batch receives only one boosting iteration. GBRL, designed explicitly for incremental updates, remains much faster.}
\label{tab:gbt_vs_libraries}
\end{table}

In the GBRL framework (\cref{fig:gbrl_diagram}) we adapt GBTs to handle the challenges of RL. To do so we reinterpret \(F_K(\state)\) as the parameterization of the policy and value function, where the observed state \(\state \in \mathcal{S}\) is the input to the ensemble. 

Specifically, at each iteration \(k\), we compute the policy gradient \cite{NIPS1999_464d828b}. As opposed to the original formulation, where the gradient is calculated with respect to parameter space, here we calculate it in function space. Consequently, we express the policy gradient as:
\begin{equation}
\nabla_{F_{k-1}} J(\pi_{F_{k-1}}) = \E_{\pi_{F_{k-1}}}[\nabla_{F_{k-1}}\log\pi_{F_{k-1}}(\action|\state) A(\state, \action)],
\end{equation}
where \(F_{k-1}\) is the current ensemble-based parametrization. 

Then, the policy gradient is projected onto a new tree and added to the ensemble:
\begin{equation}
h_k = \arg\min_{h \in \mathcal{H}} \| \nabla_{F_{k-1}}J(\pi_{F_{k-1}})  - h(\state) \|_2^2 \, .
\end{equation}
The result is an application of GBT as a functional gradient optimizer, updating both actor and critic incrementally:
\begin{equation}\label{eqn:gbrl is policy gradient}
F_{K}(\state) = F_0 + \sum_{k=1}^{K}\epsilon h_k(\state) \approx F_0 + \sum^{K}_{k=1}\epsilon\nabla_{F_{k-1}}J(\pi_{F_{k-1}}) .
 \end{equation}

\paragraph{Gradient-based Approach.} As seen in \cref{eqn:gbrl is policy gradient}, GBRL is an online learning gradient-based optimizer. In each step, a new tree is constructed to minimize the sampled loss. This learning scheme is identical to the common policy gradient methods \cite{a2c}. By leveraging optimization frameworks for gradient computation, such as PyTorch \cite{paszke2019pytorchimperativestylehighperformance}, GBRL can be integrated with most Actor-Critic algorithms and implemented within existing RL libraries. In contrast, traditional GBT frameworks are designed for offline, static datasets and require significant customizations and workarounds to function in RL. 

\begin{figure}[bt]
\centering
        \resizebox{\linewidth}{!}{
    \includegraphics{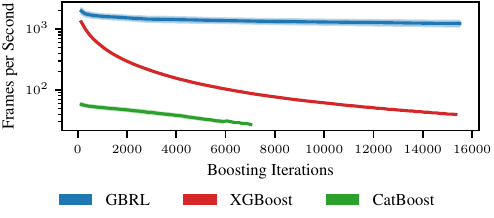}
}
\caption{\textbf{GBT library comparison, Cartpole.} CatBoost and XGBoost are intractable in RL. CatBoost's lack of GPU support for custom losses leads to low FPS and early termination.}
\label{fig:gbrl_vs_boost}
\end{figure}

\begin{figure}[bt]
\centering
        \resizebox{\linewidth}{!}{
    \includegraphics[width=\linewidth]{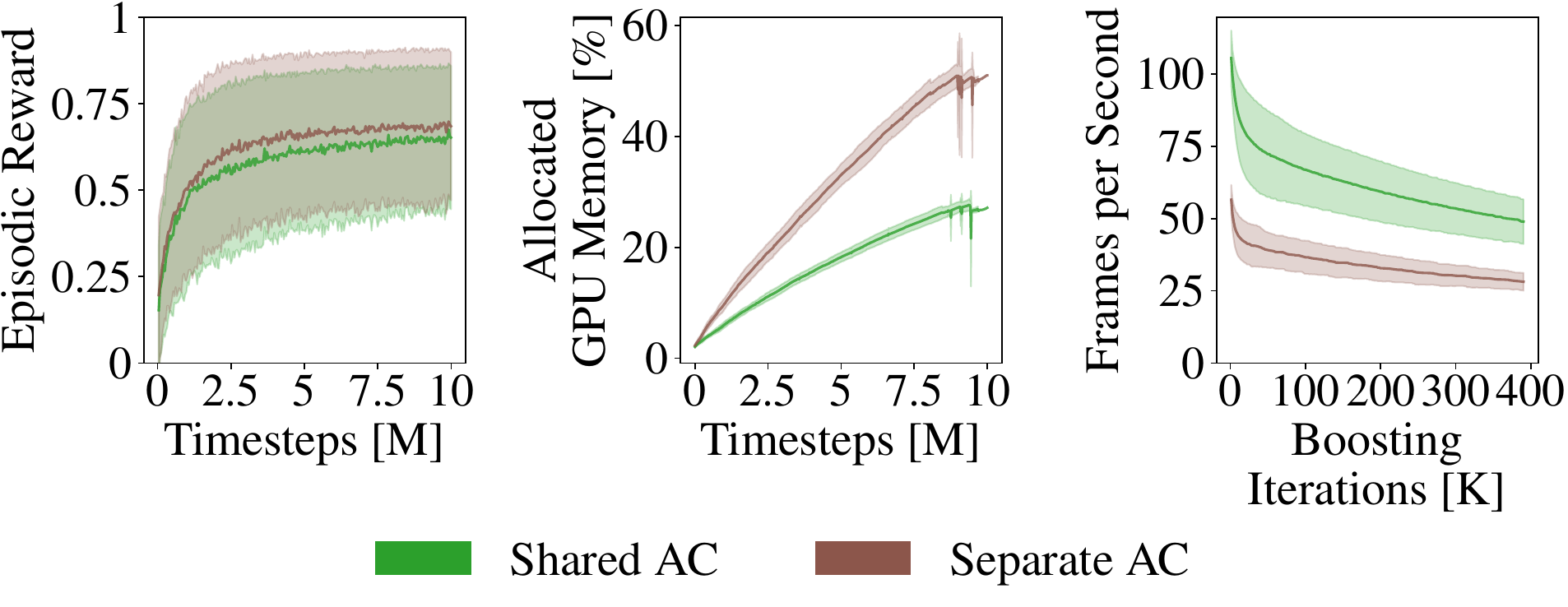}
}
    
 \caption{\textbf{Shared Actor-Critic, MiniGrid.} The shared tree structure significantly increases efficiency, without impacting the score. Aggregated results over three tasks are shown here, full per-task curves are available in Appendix (\cref{fig:ablation_env})} 
\label{fig:gbrl_efficient}
\end{figure}

\paragraph{Shared Actor-Critic Architecture.} Many RL algorithms operate with a shared Actor-Critic structure. This structure helps avoid overfitting and improves overall convergence speeds \cite{henderson2019deepreinforcementlearningmatters, andrychowicz2020mattersonpolicyreinforcementlearning}. We adopt this approach in GBRL such that each leaf produces two predictions. GBRL predicts both the policy (distribution over actions) and the value estimate. The internal structure of the tree is shared, providing a single feature representation for both objectives and significantly reduces memory and computational bottlenecks. Similar to the standard practice, we support and use separate learning rates for the policy and value. This approach enables solving two distinct objectives within a single shared structure. 

\subsection{Comparison to Traditional GBT Libraries}\label{subsec:comparison_traditional}
Before diving into the comparisons with NNs, \cref{sec: exp setup}, we first evaluated our design choices against traditional GBT libraries. While these libraries can be adapted for RL using custom loss functions and incremental learning, experiments with CatBoost and XGBoost reveal scalability issues inherent to their supervised learning design. \cref{tab:gbt_vs_libraries} shows these frameworks are significantly slower than GBRL, when used for incremental learning. Incremental learning is a fundamental property of RL, as the policy is continually changing. Moreover, when training on CartPole (\cref{fig:gbrl_vs_boost}), larger ensemble sizes lead to significant drops in training throughput. In contrast, our shared Actor-Critic architecture halves memory usage and doubles throughput without compromising policy quality (\cref{fig:gbrl_efficient}). Overall, GBRL offers a purpose-built solution that maintains GBT’s advantages while efficiently handling RL-specific needs.

\section{Experiments and Results}\label{sec: exp setup}
Our experiments aim to answer two core questions: 

 \textbf{1. GBT as an RL function approximator:} Can GBT-based AC algorithms effectively solve complex high-dimensional RL tasks? And how do they compare with NNs? 

 \textbf{2. Advantages of GBT:} Building on the GBT's success in irregular and tabular data, does its inductive bias
offer similar robustness benefits in RL -- specifically for out-of-distribution states,
noisy inputs, and spurious correlations?
 

For our experiments, we implemented a GBT-based version of PPO within Stable Baselines3 \cite{stable-baselines3}. Where available, we use standard hyperparameters, environment-specific, and normalization wrappers according to RL Baselines3 Zoo \cite{rl-zoo3}; otherwise, we optimize the hyperparameters for specific environments. For each experiment, we report the aggregated performance across five random seeds. Our training setup consists of a single NVIDIA V100 GPU. We refer the reader to the supplementary material for additional technical details, such as hyperparameters, implementation, and environment details (\cref{sec:appendix:implementation_details}), full results (\cref{appendix:results}) and full learning curves (\cref{sec:appendix:training_plots}) for all evaluated algorithms.

\subsection{Standard Environments}
\label{subsec:perf}
\paragraph{Experimental Setup.} We evaluated GBRL against NNs across three categories of RL benchmarks. First, we tested \textbf{classic RL} tasks using Classic-Control and Box2D environments from Gymnasium \cite{towers2024gymnasium}. These provided a baseline for comparison on standard benchmarks. We then evaluated both methods on more complex environments with high-dimensional vectorized representations: the \textbf{Football} domain \cite{kurach2020google}, where we employed the built-in 'Checkpoints' shaped reward, and the \textbf{Atari RAM} domain \cite{Bellemare_2013}. Finally, we assessed performance on \textbf{categorical environments}, specifically targeting the MiniGrid domain \cite{MinigridMiniworld23} -- a setting where GBTs have traditionally excelled in supervised learning. MiniGrid provides an ideal test environment through its 2D grid-world structure featuring goal-oriented tasks with discrete object interactions.

\begin{figure}[!hbt]
\centering
\centering
    \resizebox{0.9\linewidth}{!}{
    \includegraphics{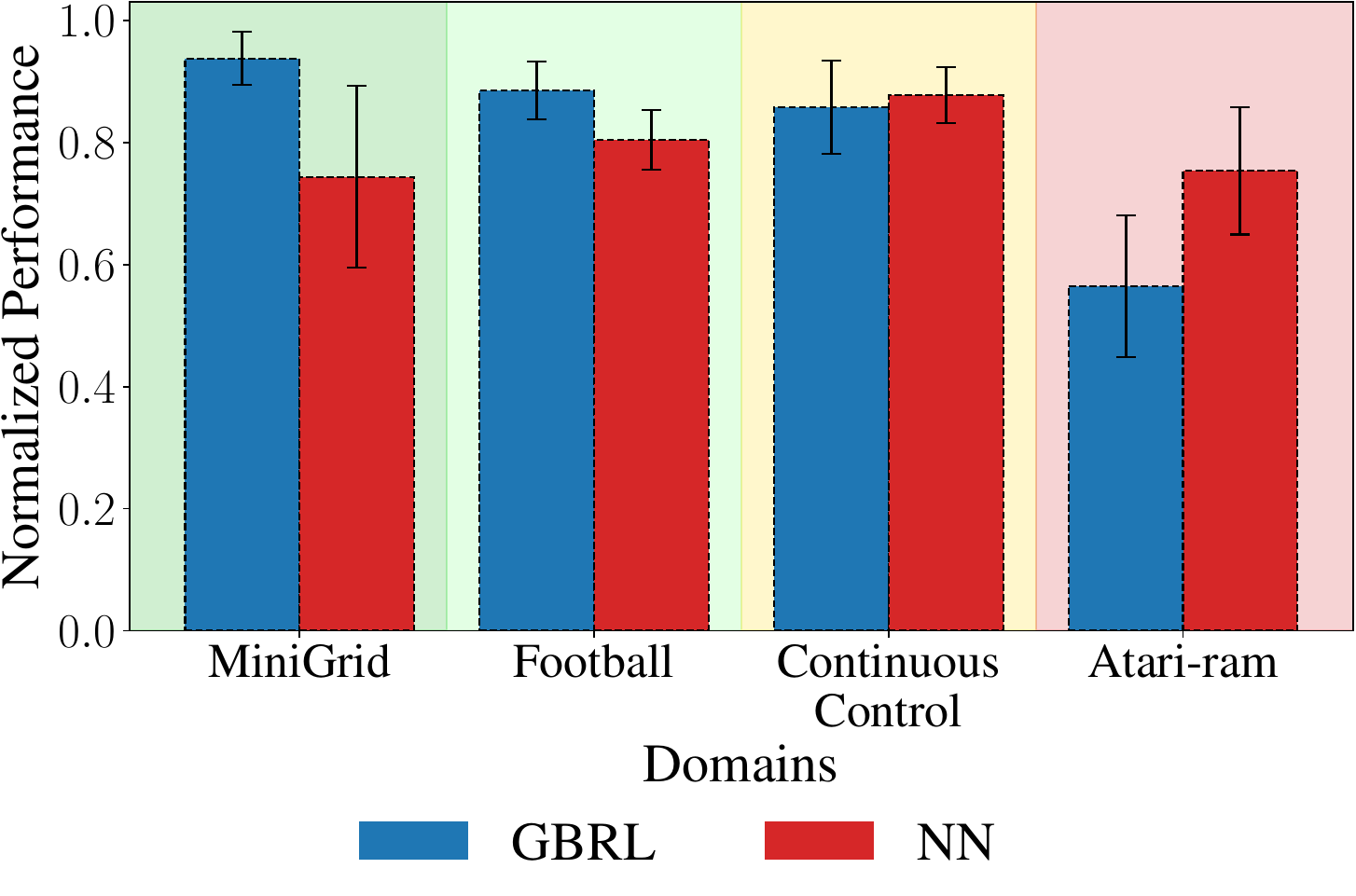}
}
    \caption{\textbf{GBRL vs NN in standard environments (PPO).} Aggregated mean and standard deviation of the normalized average reward for the final 100 episodes.} 

    \label{fig:best_algo}
\end{figure}

\paragraph{Results.} We report the cumulative non-discounted reward, averaged across the last 100 episodes in \cref{fig:best_algo}. For a simple visual comparison between GBRL and NNs, we report the normalized score: 
\(\reward_{\text{norm}} = \frac{\reward_{\text{GBRL}} - \reward_{\text{NN}}}{\reward_{\text{max}\{\text{NN}, \text{GBRL}\}}- \reward_{\text{min}\{\text{NN}, \text{GBRL}\}}} \). 

Decision trees, at their core, are if-else structures that excel in handling numerical and categorical data. In MiniGrid, which provides such environments, GBRL significantly outperformed NN. For example, in tasks like PutNear, FourRooms, and Fetch, GBRL consistently achieved higher rewards, highlighting its ability to exploit the structured nature of the environment effectively. This performance demonstrates that GBRL is particularly well-suited for problems where data can be neatly partitioned, aligning with GBT’s inherent design strengths. This suggests that structured environments represent an ideal use case for GBRL.

An additional domain characterized by structured representations is the Football domain, where we used a shaped reward. Here, features such as player positions and ball locations were designed to represent identifiable, interpretable information. In this domain, GBRL outperformed NN across most scenarios and exhibited equivalent performance in the rest. The structured nature of the Football domain aligns well with the strengths of GBT, effectively partitioning the feature space. Interestingly, while both Football and MiniGrid are structured, the Football domain is not sparse. Hence, GBRL’s strong performance in Football underscores its robustness and capacity to generalize well in structured, high-dimensional environments.

Unlike Football, the Atari RAM domain offers unstructured, high-dimensional representations by exposing raw system memory states. These flattened views lack explicit semantic structure, making them challenging for GBRL, which relies on single-feature splits. As a result, GBRL underperformed compared to NN in most Atari RAM tasks. This result highlights an important limitation of decision-tree-based approaches. They struggle to model implicit feature interactions, an inherent characteristic of unstructured data.

Finally, classic control tasks are simpler environments that rely on low-dimensional state spaces. In these tasks, GBRL demonstrated performance comparable to that of the NN counterpart. This suggests that decision tree-based models can match NNs in solving tasks where feature interactions are limited and the optimization landscape is smooth.

Overall, these comparisons indicate that GBT can serve as a strong function approximator in RL, matching or surpassing NN's performance in certain environments.

\subsection{Challenge Environments}
Having established GBRL's general capabilities, we now examine scenarios designed to test inherent properties of decision trees. These experiments focus on challenging conditions---sparse rewards, misleading correlations, and input perturbations---where NNs typically struggle.

\subsubsection{Signal Dilution}
Signal dilution occurs when rare events, such as sparse rewards, are statistically overwhelmed by more common non-event data during training \cite{shyalika2024comprehensive,zhao2018framework,he2021weighting,tessler2017deep}. In NNs with fixed architectures, shared weights inherently prioritize patterns from dominant classes or frequent signals, causing rare events to be "averaged out" or under-represented in learned representations. This phenomenon is particularly problematic in RL and classification tasks with imbalanced datasets, where the model fails to retain or react to critical but infrequent signals.

\paragraph{Experimental Setup.} We compare GBRL with NNs in a linear equation variable isolation environment with sparse rewards, inspired by \cite{poesia2021contrastivereinforcementlearningsymbolic}. The agent observes the coefficients $(a,b,c)$ of a linear equation $ax + b = c$, and needs to isolate \(x\), receiving a large reward only upon completion. Actions are multi-discrete, involving the selection of an operation, a digit \(\in [0,9]\), and its sign. The optimal policy solves the simple task in 2 steps. We also evaluate two more complex variations with an additional variable \(y\), where the optimal policy requires 3–4 steps. For these, we introduce intermediate goals with small rewards to guide the agent. 

\begin{figure}[bt]
\centering
    \includegraphics[width=\linewidth]{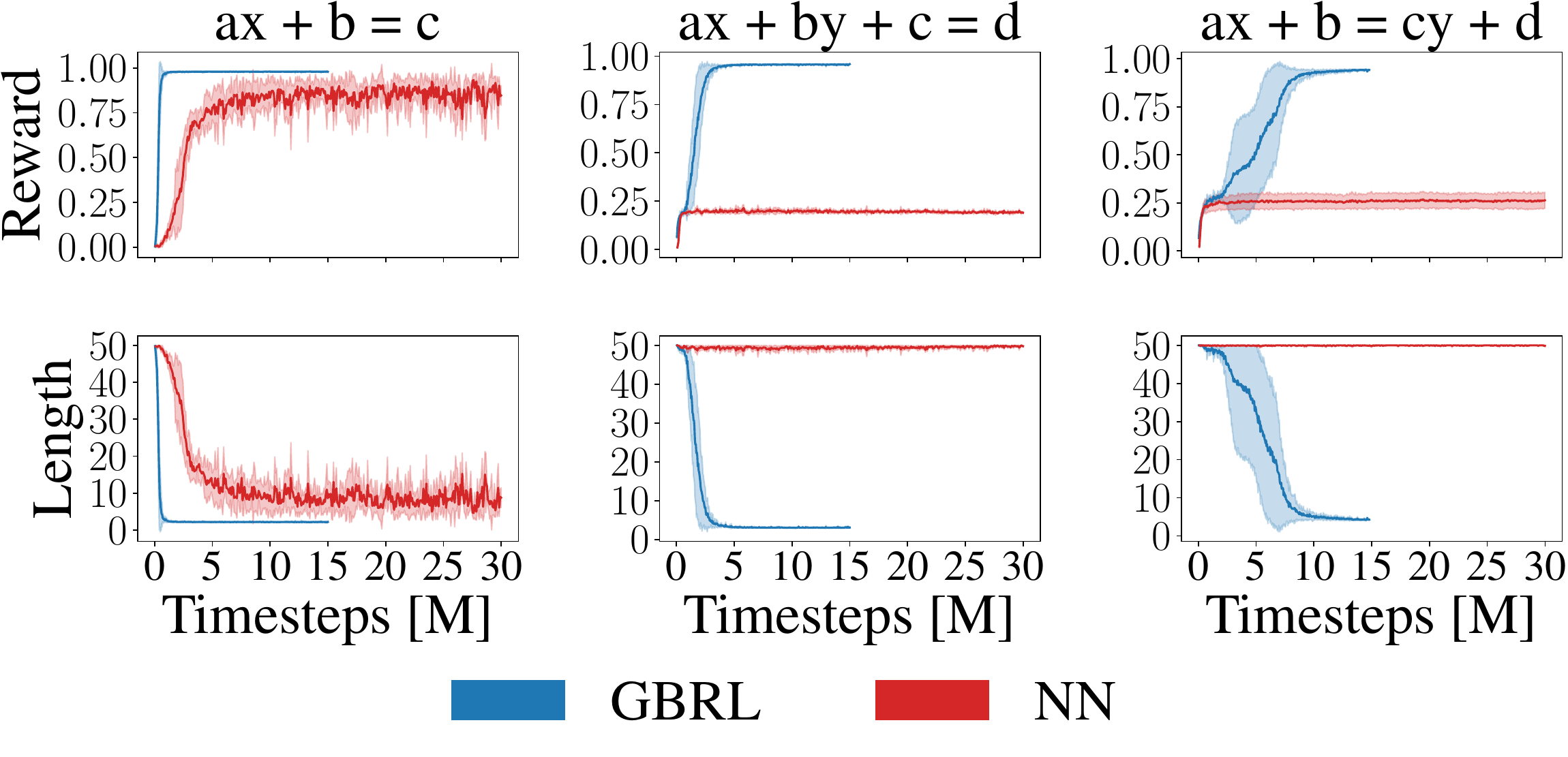}
\caption{\textbf{Signal dilution, variable isolation task.} Mean and standard deviation of the average episodic reward during training. GBRL was trained for 15M training steps and NN for 30M. Episodes are terminated after 50 steps if not solved.}
\label{fig:equation_results}
\end{figure}

\paragraph{Results.} While simple for humans, these tasks remain challenging for NNs due to signal dilution (\cref{fig:equation_results}). Although they can make some progress in the simple task ($ax+b=c$), they fail when the equation complexity increases, even when given intermediate rewards. In contrast, GBRL converges to and retains a stable (and optimal) policy throughout the process. GBTs naturally segment the decision space into explicit rule-based splits, preventing signal dilution while correctly preserving the rare but crucial reward signals. Consequently, we see that GBRL can preserve and exploit rare but critical reward signals—evidence that GBT-based methods are inherently less prone to signal dilution.

\subsubsection{Spurious Correlations}
\label{subsec:spurious}
\paragraph{Experimental Setup.} The agent is placed in a grid environment with three balls (red, green, and blue), and a red box. The ball locations are randomized. The agent observes an objective specifying which ball to pick up. We experiment with three box-placement configurations: adjacent to the target ball, adjacent to one of the non-target balls, and without a box in the room. This task tests whether the policy overfits spuriously correlated features. We then analyze the results using SHAP \cite{shap}, a method based on Shapley values \cite{shapley:book1952} that measures the contribution of each feature to the final prediction of a model. 

\begin{figure}[bt]
\centering
    \resizebox{\linewidth}{!}{
    \includegraphics{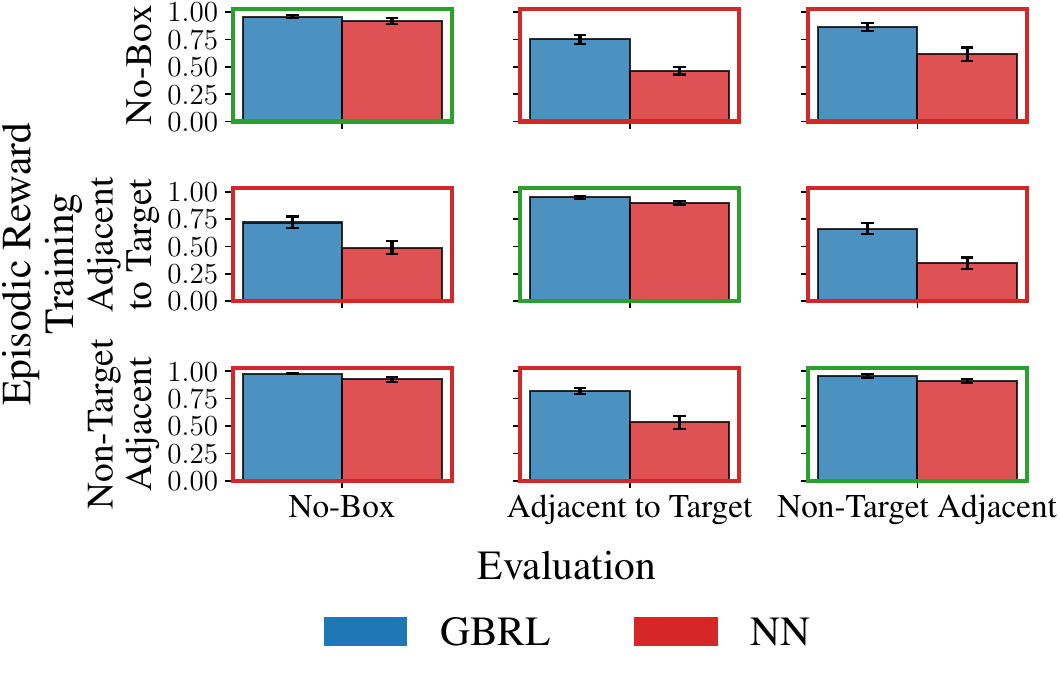}
}
\caption{\textbf{Spurious correlations, MiniGrid.} Mean and standard deviation of the episodic reward. Plots outlined in \textcolor{tabgreen}{green} are in-distribution, while plots outlined in \textcolor{tabred}{red} are out-of-distribution.}
\label{fig:minigrid_confusion}
\end{figure}

\paragraph{Results.} GBRL consistently outperforms NNs in all scenarios (\cref{fig:minigrid_confusion}), demonstrating its ability to ignore irrelevant distractions, such as the box, and reliably reach the target regardless of its placement.

\begin{figure}[bt]
\centering
\begin{subfigure}{\linewidth}
\centering
    \includegraphics[width=0.75\linewidth]{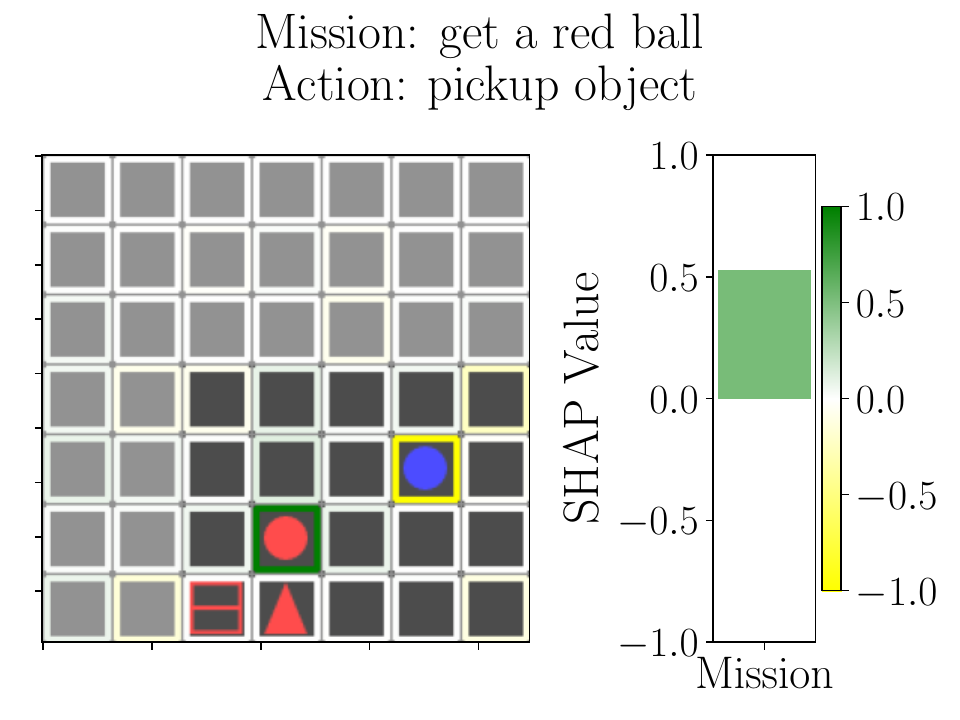}
    \caption{\textbf{GBRL SHAP.}} 
    \label{fig:gbrl shap}
\end{subfigure}
\begin{subfigure}{\linewidth}
\centering
    \includegraphics[width=0.75\linewidth]{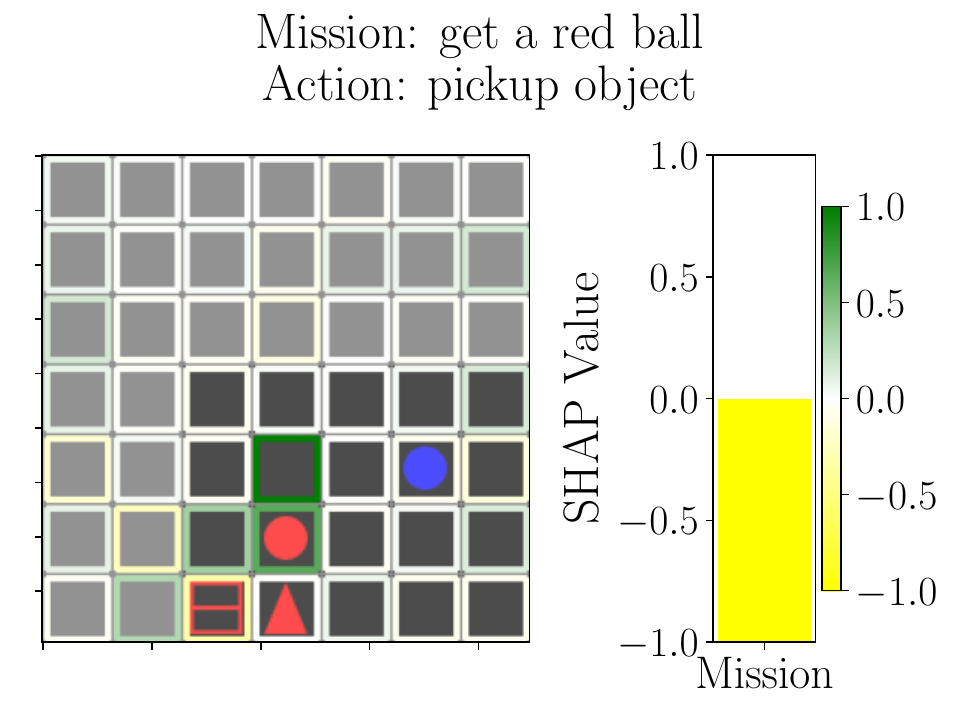}
    \caption{\textbf{NN SHAP.}} 
    \label{fig:nn shap}
\end{subfigure}
    
\caption{\textbf{Spurious correlations revealed by feature importance in MiniGrid.} The visualization shows normalized SHAP values for the concluding action in a successful evaluation run, revealing the magnitude and direction of each feature’s impact on model predictions. In this scenario, a red box was positioned next to the target (red ball), and both models converged to a successful policy. GBRL prioritizes the mission, ignores the red box, and contrasts the blue ball. In contrast, the NN focuses on the cells surrounding the target, while ignoring the blue ball and contrasting the box and the mission.}
\label{fig:minigrid_ood_shap}
\end{figure}

To understand GBRL's performance, we analyzed the SHAP values for the agent's surroundings and the mission feature when predicting the logit for picking up the target object (\cref{fig:minigrid_ood_shap}). GBRL assigns the strongest positive SHAP value to the target object, followed by the mission string, aligning perfectly with the goal. Non-target objects receive the most negative SHAP values, suppressing distractions and reinforcing goal-directed behavior. In contrast, NNs assign negative SHAP values to the mission feature and rely heavily on the grid surrounding the target object. This indicates a reliance on spurious correlations and a lack of goal-oriented strategies.

These findings underscore GBRL’s ability to focus on task-relevant features and ignore deceptive cues, reinforcing our hypothesis that GBT’s inductive bias provides robustness against spurious correlations. Next, we consider how this robustness extends to other forms of state perturbations.

\subsubsection{Robustness to State Perturbations}
When encountering an out-of-distribution (OOD) state, a GBT model will always output a result within the same range seen in training. In contrast, NNs may extrapolate and exhibit unexpected behavior \cite{xu2021neuralnetworksextrapolatefeedforward, meinke2020neuralnetworksprovablyknow, ulmer2021knowlimitsuncertaintyestimation}. We consider three tasks. First, robustness to irrelevant information. Second, robustness to random noise. And finally, robustness to missing features. These experiments aim to test the inherent robustness properties of GBTs when encountering different types of noise during training.

\paragraph{Irrelevant Information.} We consider a similar fetch experiment as before (\cref{subsec:spurious}). Now, a new object is placed in the room at a random location. This object is not part of the mission and only interferes with navigation. The interfering object differs from the task objects in shape, color, or both.

\begin{figure}[bt]
\centering
    \resizebox{\linewidth}{!}{
    \includegraphics{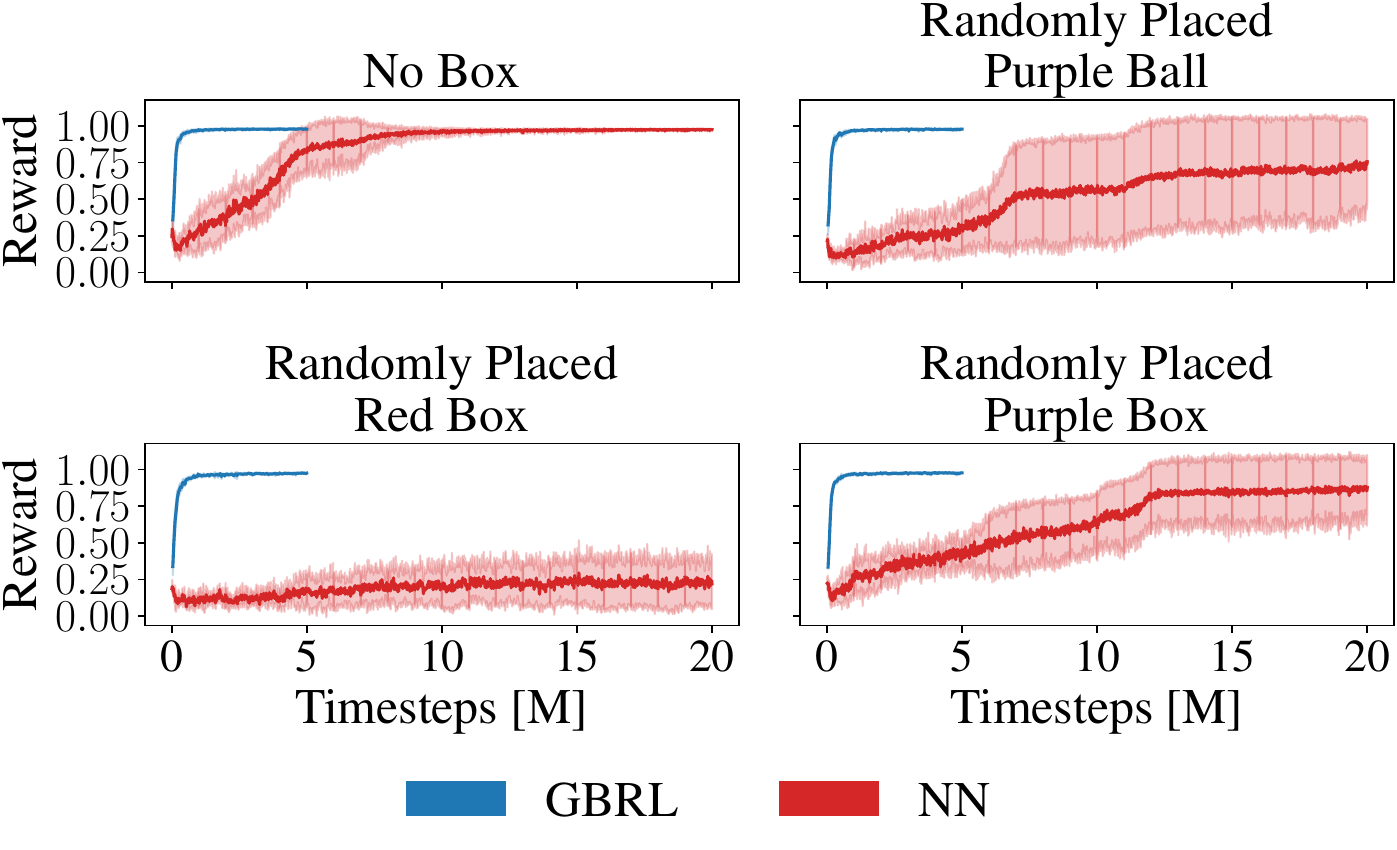}
}
\caption{\textbf{Robustness to irrelevant objects, MiniGrid.} Mean and standard deviation of the average episodic reward during training. GBRL was trained for 5M training steps, and NN for 20M.}
\label{fig:minigrid_random}
\end{figure}

\cref{fig:minigrid_random} illustrates how introducing a distractor impacts policy performance. 
When the environment is free of distracting objects, NNs are able to solve the task, converging to a stable and ideal solution. Surprisingly, although the purple-box object shares no features with the task objects (blue, green, and red balls), NNs exhibit a drop in performance but are able to solve the task in a subset of seeds. This performance degradation increases when the confusing object shares more features with the goal objects, such as purple-\textit{ball} and \textit{red}-box. In contrast, GBRL successfully and efficiently solves the task, regardless of the additional irrelevant object placed in the scene.

\paragraph{Random Noise.} In the classic-control environments (\cref{subsec:perf}), we train agents in a noiseless setting. Then, to evaluate the robustness of the resulting policies, we add random zero-centered Gaussian noise to the input states as defined by: \(\state_t = \state_t + \epsilon|\state_t|\). 

\begin{figure}[bt]
\centering
    \resizebox{\linewidth}{!}{
    \includegraphics{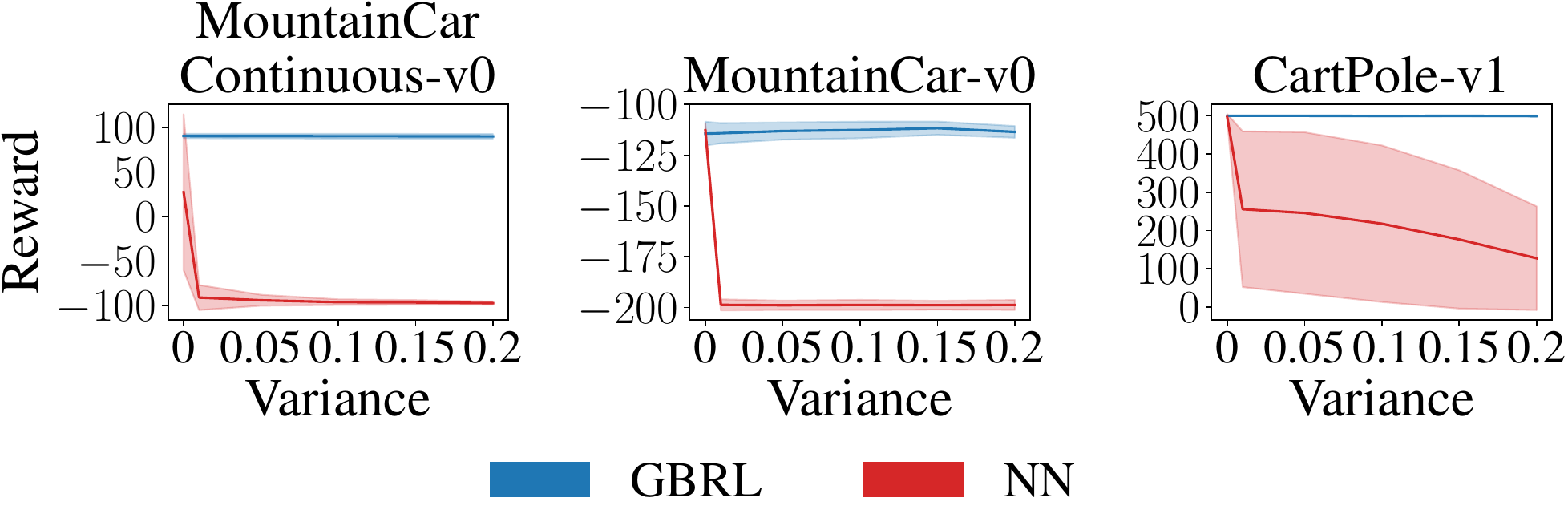}
}
\caption{\textbf{Robustness to input perturbations, classic-control.} Models were trained without perturbations and then evaluated on noisy observations. At each timestep, the observation was perturbed following \(\state_t = \state_t +\epsilon|\state_t|\), where \(\epsilon \sim \mathcal{N}(0, \text{var}) \). We report the mean and standard deviation of episodic reward across 100 evaluation episodes for each \(\text{Variance} \in [0, 0.2]\).}
\label{fig:gym_noise} 
\end{figure}

The results show that these characteristics translate to practical outcomes (\cref{fig:gym_noise}). In classic-control environments, a trained GBT-based policy is much more robust to small variations in the state space compared to NNs. We attribute this performance to how GBTs partition the input space. This partitioning occurs on hard thresholds that are split according to single features. Although small perturbations may be sufficient to completely change decision outcomes in NNs \cite{certifiedadversarialrobustnessdeep, carlini2017adversarialexampleseasilydetected}, GBRL exhibits robustness to random additive noise.

\paragraph{Missing Features.} We modify scenarios in the football domain (\cref{subsec:perf}) to support the random drop of a player from the opposing team. 
Policies are trained on both the original task and a modified variant. We then evaluate model robustness by testing how a policy trained against \(N\) random players performs when faced with \(N+1\), or how a policy trained against a full team behaves when a random opponent is removed.

\begin{figure}[bt]
\centering
    \resizebox{\linewidth}{!}{
    \includegraphics{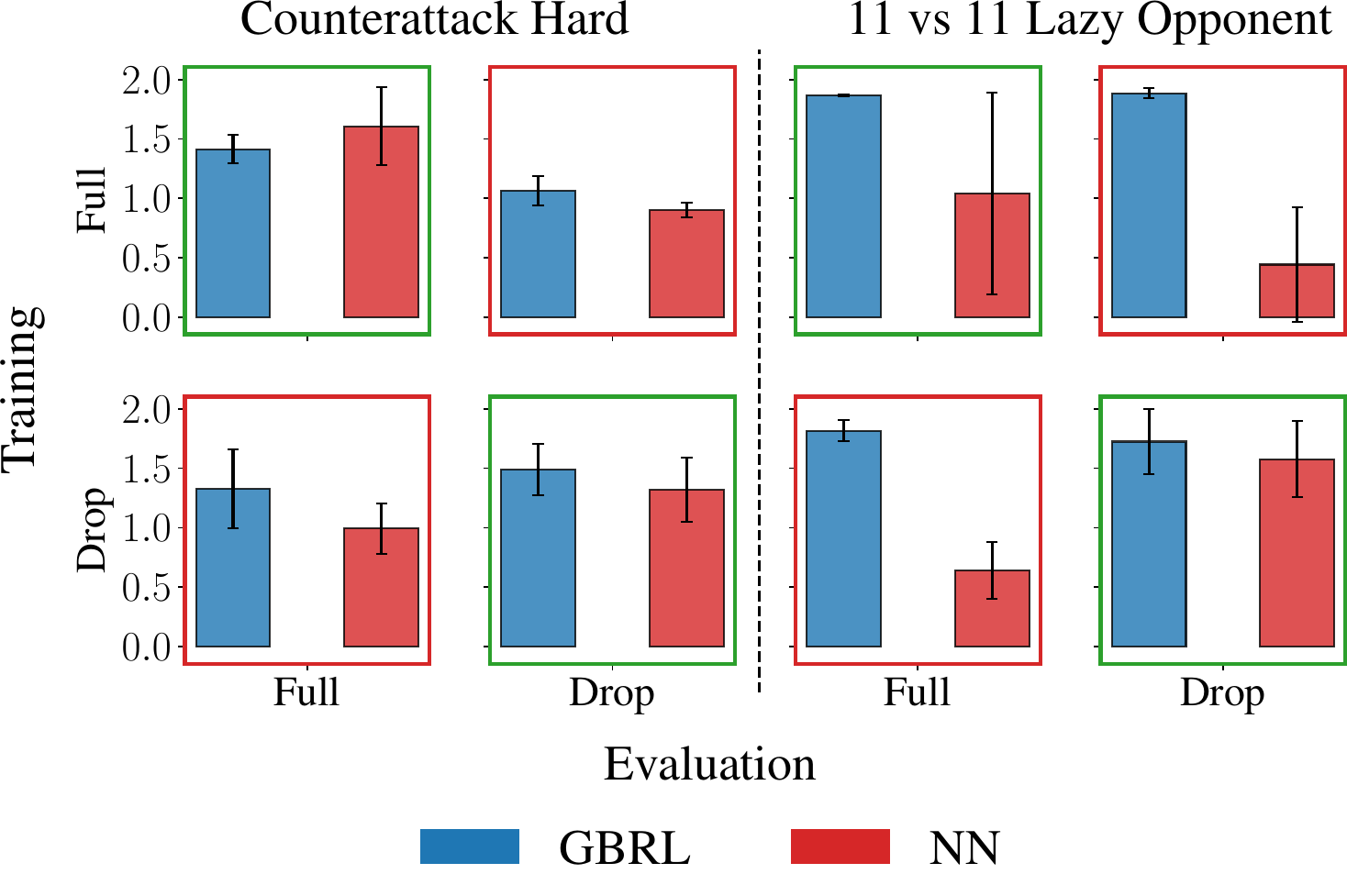}
}
\caption{\textbf{Out-of-distribution evaluation, football.} Mean and standard deviation of episodic reward across 100 evaluation episodes. Plots outlined in \textcolor{tabgreen}{green} are in-distribution, while plots outlined in \textcolor{tabred}{red} are out-of-distribution.}
\label{fig:football_ood}
\end{figure}

The results show that GBRL outperforms NNs in OOD tasks (\cref{fig:football_ood}), particularly in the 11 vs 11 academy scenario. Notably, GBRL maintains performance when evaluated against 10 opponents and can solve 11 vs. 11 even when trained on 10. In contrast, NNs struggle in all OOD cases. This suggests GBRL better isolates features that are essential to solving the task.

\section{Ablation Study}
In this section, we examine the sensitivity of GBRL to key hyperparameters. We systematically changed one hyperparameter at a time while keeping the others fixed.
For each experiment, we trained an agent in the CartPole-v1 environment for 1M timesteps using GBT-based PPO.

\begin{figure}[bt]
\centering
    \includegraphics[width=\linewidth]{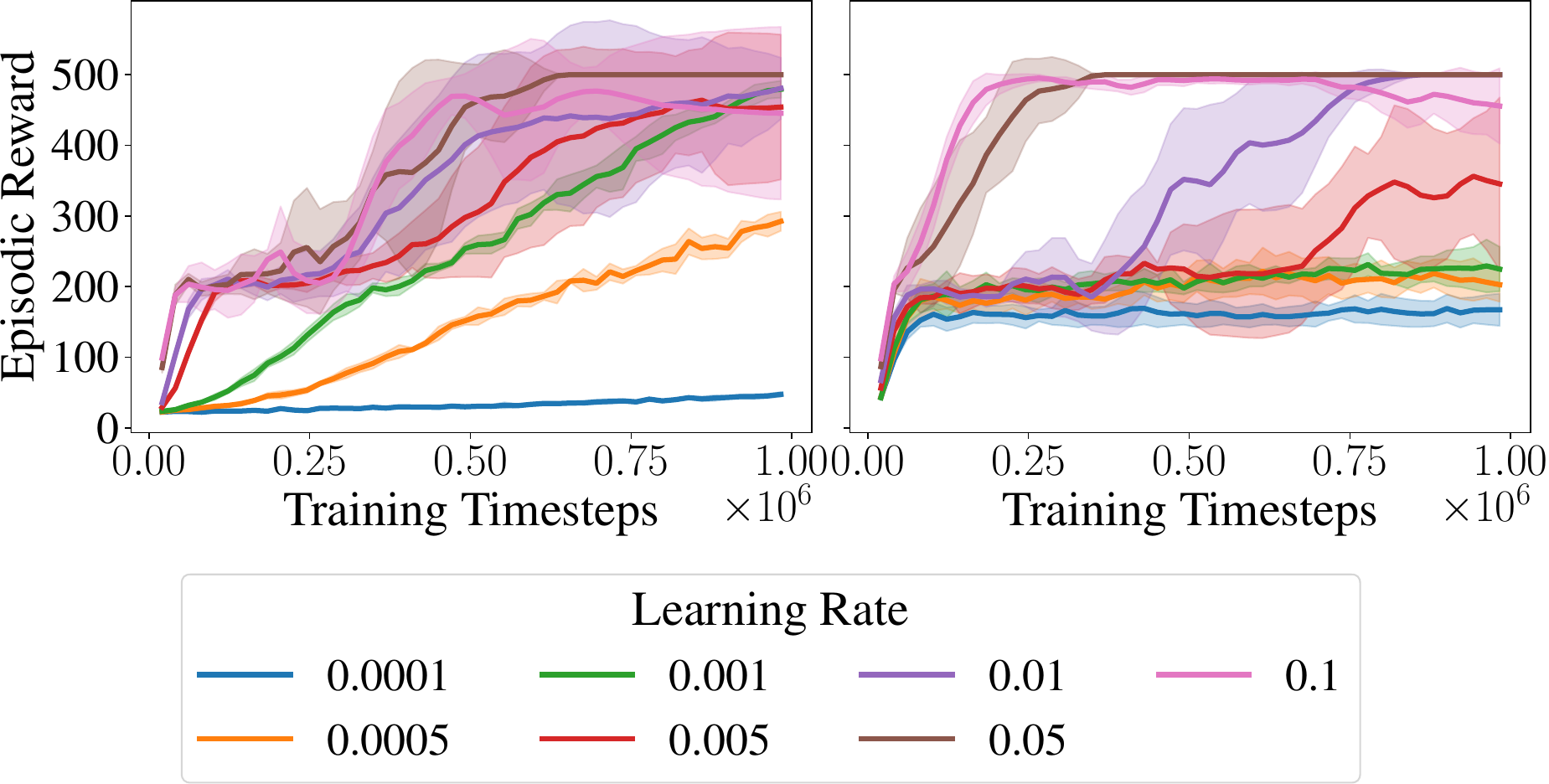}
\caption{\textbf{Ablation: Learning rate variation across policy (left) and value function (right) components}. Mean and
standard deviation of the average episodic reward during training. Each subplot shows the effect of varying the learning rate for one component (policy or value function) while keeping the other fixed.}
    \label{fig:ablation_lr}
\end{figure}

\paragraph{\textbf{Learning Rate Variation Across Policy and Value Function Components}.} We investigated the impact of the learning rate for both the policy and value function components. In these experiments, we held one learning rate constant while varying the other. Consistent with observations in NN training, we found that excessively high learning rates destabilized the learning process, whereas overly low rates led to significantly slower convergence (as illustrated in \cref{fig:ablation_lr}).

\begin{figure}[bt]
\centering
    \includegraphics[width=\linewidth]{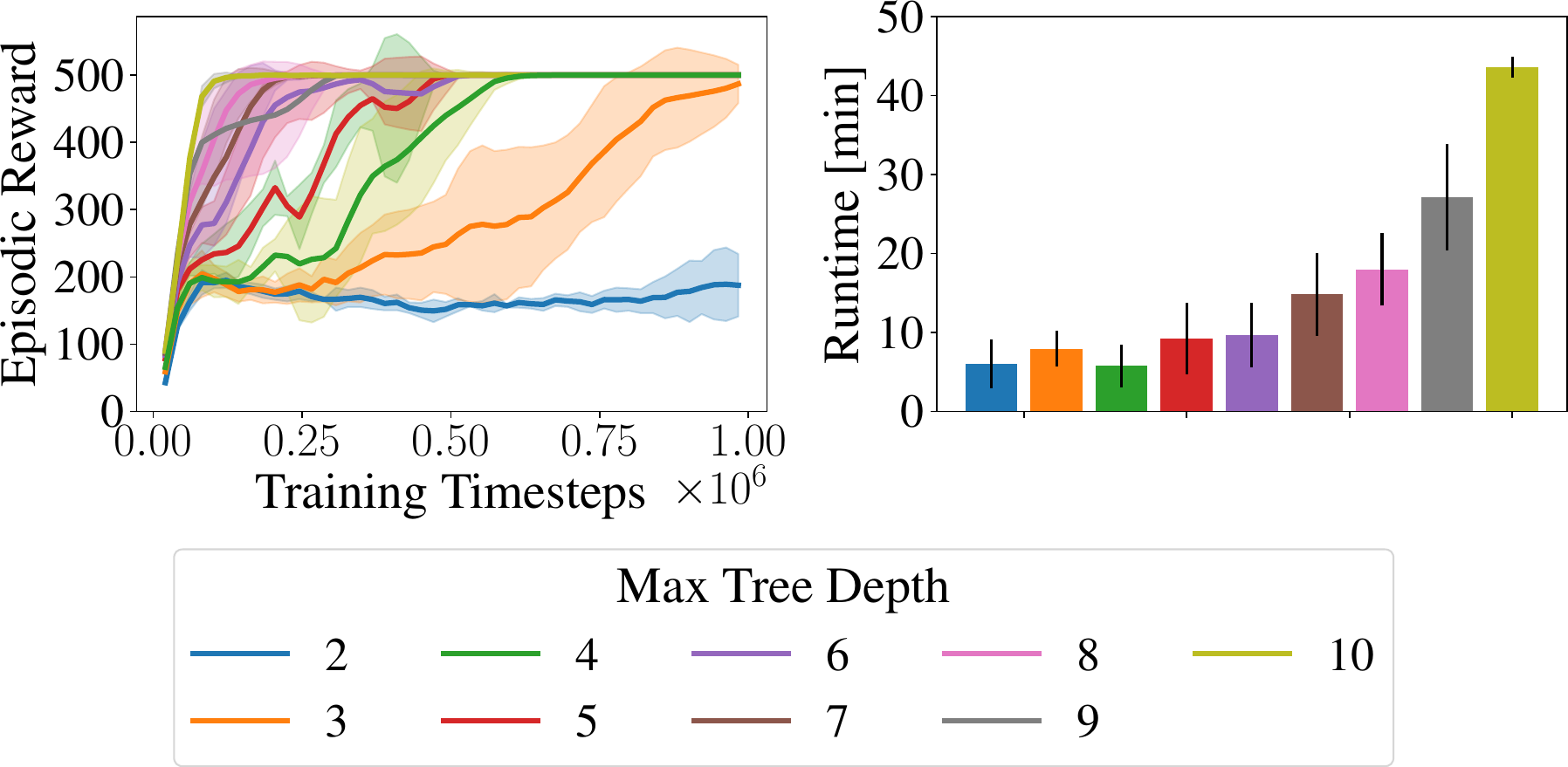}
\caption{\textbf{Ablation: Effect of tree depth on convergence}. We varied the maximum depth of the decision trees used in the gradient-boosted models and evaluated their impact on learning dynamics. The plot shows the mean and standard deviation of the average episodic reward over time and the total runtime.} 
    \label{fig:ablation_depth}
\end{figure}

\paragraph{\textbf{Limitations of Tree Depth}.} We varied the maximum depth of the decision trees built at each training iteration and evaluated their impact on
learning dynamics. \cref{fig:ablation_depth} demonstrates that deeper trees yield more accurate gradient approximations, leading to faster convergence, but also incur a higher computational cost per training step. In the paper, we selected a maximum tree depth of 4, balancing convergence speed and wall-clock time.

\begin{figure}[bt]
\centering
    \includegraphics[width=0.95\linewidth]{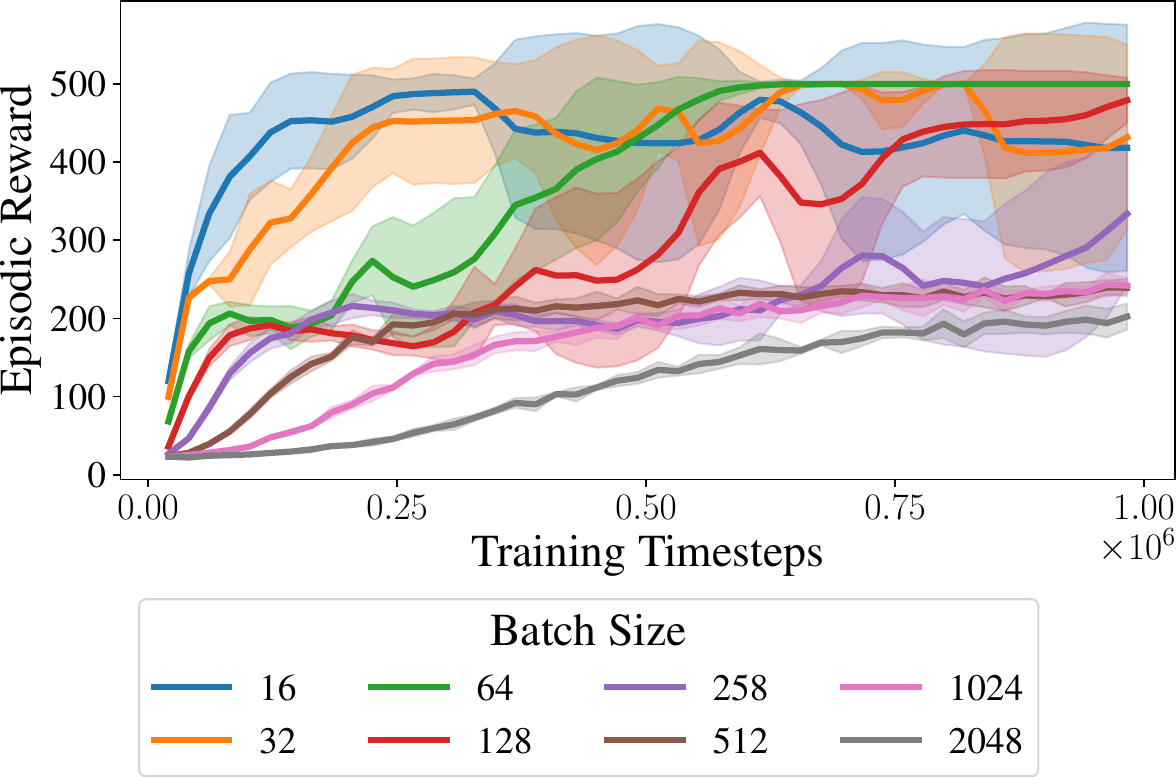}
\caption{\textbf{Ablation: Batch size impact on convergence}. Mean and
standard deviation of the average episodic reward during training. We fix the rollout length to 2048 and vary the batch size. Smaller batches improve early learning but suffer from instability, while larger batches stabilize training at the cost of slower convergence.}
    \label{fig:ablation_batch}
\end{figure}

\paragraph{\textbf{Batch Size Impact on Convergence}.} We analyzed the impact of the batch size on performance. In \cref{fig:ablation_batch}, we observe that batch size significantly impacts convergence. Specifically, smaller batches result in GBRL building more trees per rollout, improving adaptability. However, smaller batches also lead to noisier gradient estimates as a result of limited samples per constructed tree. Conversely, larger batches stabilize training by reducing variance through averaging within leaves, as more samples are utilized to construct each tree but build fewer trees per rollout. Hence, both excessively small and large batch sizes negatively impact performance.

Overall, these ablations highlight that GBRL's performance is sensitive to training frequency, tree depth, and batch size—each of which governs the trade-off between convergence speed and stability. While this section focused on general hyperparameters, we further investigate how these trade-offs manifest across different RL algorithms in \cref{sec:appendix:algo_diff}. 

\section{Limitations and Future Directions}
GBRL exhibits competitive performance and surprising robustness capabilities. However, despite these benefits, we highlight two important limitations.
\textbf{(i) Continuous generation of trees.}  As the policy improves through numerous updates, the size of the ensemble increases. This unbounded growth has implications for memory usage, computational efficiency, and the feasibility of online real-time adaptation. 
Future work may leverage tree redundancy to produce a more efficient and compact policy. For example, strategies for tree pruning, ensemble compression, or dynamically managing ensemble size. 
\textbf{ (ii) Off-policy continuous control.} Methods like DDPG \cite{ddpg} and SAC \cite{haarnoja2018soft} update the policy by differentiating the Q-estimator. As GBTs are not differentiable, new solutions are needed to incorporate them into these algorithms. 

\section{Conclusion}
Historically, RL practitioners have relied on tabular, linear, and NN-based function approximators. While GBT evolved as a widely successful tool in supervised learning, it has been absent from this toolbox. In this work, we introduced a method for effectively integrating GBT into RL. We demonstrated domains where GBT excels compared to NNs, in addition to analyzing its various inherent characteristics.

We observe that the ideal method depends on the task characteristics. Tabular and linear approaches are suitable for small state spaces or simple mappings, while NNs excel when tackling complex relationships in unstructured data. In comparison, GBT thrives in complex, yet structured environments. In such cases, GBRL's advantage is highlighted, reflecting GBTs known benefits from supervised learning.

A crucial component of GBRL is our efficient adaptation of GBT for Actor-Critic methods, which allows the simultaneous optimization of distinct objectives. We optimized this approach for large-scale ensembles using GPU acceleration (CUDA). Furthermore, GBRL integrates seamlessly with existing RL libraries, promoting ease of use and adoption.

GBRL is a step toward well-suited solutions for real-world tasks with structured data.

\section*{Impact Statement}

This paper presents work whose goal is to advance the field of 
Machine Learning. There are many potential societal consequences 
of our work, none of which we feel must be specifically highlighted here.

\bibliography{icml}

@article{shyalika2024comprehensive,
author = {Shyalika, Chathurangi and Wickramarachchi, Ruwan and Sheth, Amit P.},
title = {A Comprehensive Survey on Rare Event Prediction},
year = {2024},
issue_date = {March 2025},
publisher = {Association for Computing Machinery},
address = {New York, NY, USA},
volume = {57},
number = {3},
issn = {0360-0300},
url = {https://doi.org/10.1145/3699955},
doi = {10.1145/3699955},
journal = {ACM Comput. Surv.},
month = nov,
articleno = {70},
numpages = {39}
}

@article{he2021weighting,
  title={Weighting methods for rare event identification from imbalanced datasets},
  author={He, Jia and Cheng, Maggie X},
  journal={Frontiers in big Data},
  volume={4},
  pages={715320},
  year={2021},
  publisher={Frontiers Media SA},
  url={https://www.frontiersin.org/journals/big-data/articles/10.3389/fdata.2021.715320},
 doi={10.3389/fdata.2021.715320},
}

@article{zhao2018framework,
author = {Zhao, Yang and Wong, Zoie Shui-Yee and Tsui, Kwok Leung},
title = {A Framework of Rebalancing Imbalanced Healthcare Data for Rare Events’ Classification: A Case of Look-Alike Sound-Alike Mix-Up Incident Detection},
journal = {Journal of Healthcare Engineering},
volume = {2018},
number = {1},
pages = {6275435},
doi = {https://doi.org/10.1155/2018/6275435},
url = {https://onlinelibrary.wiley.com/doi/abs/10.1155/2018/6275435},
eprint = {https://onlinelibrary.wiley.com/doi/pdf/10.1155/2018/6275435},
year = {2018}
}

@article{10.1214/aos/1013203451,
    author = {Jerome H. Friedman},
    doi = {10.1214/aos/1013203451},
    journal = {The Annals of Statistics},
    number = {5},
    pages = {1189 -- 1232},
    publisher = {Institute of Mathematical Statistics},
    title = {{Greedy function approximation: A gradient boosting machine.}},
    url = {https://doi.org/10.1214/aos/1013203451},
    volume = {29},
    year = {2001}
}

@manual{nvidia_cuda,
  title        = {CUDA Toolkit Documentation},
  author       = {NVIDIA},
  year         = {2025},
  note         = {Version 12.1},
  url          = {https://developer.nvidia.com/cuda-toolkit}
}

@inproceedings{NIPS1999_464d828b,
    author = {Sutton, Richard S and McAllester, David and Singh, Satinder and Mansour, Yishay},
    booktitle = {Advances in Neural Information Processing Systems},
    editor = {S. Solla and T. Leen and K. M\"{u}ller},
    pages = {},
    publisher = {MIT Press},
    title = {Policy Gradient Methods for Reinforcement Learning with Function Approximation},
    url = {https://proceedings.neurips.cc/paper_files/paper/1999/file/464d828b85b0bed98e80ade0a5c43b0f-Paper.pdf},
    volume = {12},
    year = {1999}
}

@InProceedings{haarnoja2018soft,
  title = 	 {Soft Actor-Critic: Off-Policy Maximum Entropy Deep Reinforcement Learning with a Stochastic Actor},
  author =       {Haarnoja, Tuomas and Zhou, Aurick and Abbeel, Pieter and Levine, Sergey},
  booktitle = 	 {Proceedings of the 35th International Conference on Machine Learning},
  pages = 	 {1861--1870},
  year = 	 {2018},
  editor = 	 {Dy, Jennifer and Krause, Andreas},
  volume = 	 {80},
  series = 	 {Proceedings of Machine Learning Research},
  month = 	 {10--15 Jul},
  publisher =    {PMLR},
  url = 	 {https://proceedings.mlr.press/v80/haarnoja18b.html},
}

@misc{ppo,
    doi = {10.48550/ARXIV.1707.06347},
    url = {https://arxiv.org/abs/1707.06347},
    author = {Schulman, John and Wolski, Filip and Dhariwal, Prafulla and Radford, Alec and Klimov, Oleg},
    title = {Proximal Policy Optimization Algorithms},
    publisher = {arXiv},
    year = {2017},
    copyright = {arXiv.org perpetual, non-exclusive license}
}

@InProceedings{a2c,
  title = 	 {Asynchronous Methods for Deep Reinforcement Learning},
  author = 	 {Mnih, Volodymyr and Badia, Adria Puigdomenech and Mirza, Mehdi and Graves, Alex and Lillicrap, Timothy and Harley, Tim and Silver, David and Kavukcuoglu, Koray},
  booktitle = 	 {Proceedings of The 33rd International Conference on Machine Learning},
  pages = 	 {1928--1937},
  year = 	 {2016},
  editor = 	 {Balcan, Maria Florina and Weinberger, Kilian Q.},
  volume = 	 {48},
  series = 	 {Proceedings of Machine Learning Research},
  address = 	 {New York, New York, USA},
  month = 	 {20--22 Jun},
  publisher =    {PMLR},
  url = 	 {https://proceedings.mlr.press/v48/mniha16.html},
}

@inproceedings{nonparametricrl,
    author = {Kersting, Kristian and Driessens, Kurt},
    title = {Non-parametric policy gradients: a unified treatment of propositional and relational domains},
    year = {2008},
    isbn = {9781605582054},
    publisher = {Association for Computing Machinery},
    address = {New York, NY, USA},
    url = {https://doi.org/10.1145/1390156.1390214},
    doi = {10.1145/1390156.1390214},
    booktitle = {Proceedings of the 25th International Conference on Machine Learning},
    pages = {456–463},
    numpages = {8},
    location = {Helsinki, Finland},
    series = {ICML '08}
}

@inproceedings{functional_gd,
    author = {Mason, Llew and Baxter, Jonathan and Bartlett, Peter and Frean, Marcus},
    booktitle = {Advances in Neural Information Processing Systems},
    editor = {S. Solla and T. Leen and K. M\"{u}ller},
    pages = {},
    publisher = {MIT Press},
    title = {Boosting Algorithms as Gradient Descent},
    url = {https://proceedings.neurips.cc/paper_files/paper/1999/file/96a93ba89a5b5c6c226e49b88973f46e-Paper.pdf},
    volume = {12},
    year = {1999}
}

@misc{awr,
      title={Advantage-Weighted Regression: Simple and Scalable Off-Policy Reinforcement Learning}, 
      author={Xue Bin Peng and Aviral Kumar and Grace Zhang and Sergey Levine},
      year={2019},
      eprint={1910.00177},
      archivePrefix={arXiv},
      primaryClass={cs.LG},
      url={https://arxiv.org/abs/1910.00177}, 
}

@inproceedings{xgboost,
    series={KDD ’16},
    title={XGBoost: A Scalable Tree Boosting System},
    url={http://dx.doi.org/10.1145/2939672.2939785},
    doi={10.1145/2939672.2939785},
    booktitle={Proceedings of the 22nd ACM SIGKDD International Conference on Knowledge Discovery and Data Mining},
    publisher={ACM},
    author={Chen, Tianqi and Guestrin, Carlos},
    year={2016},
    month=aug,
    collection={KDD ’16}
}

@inproceedings{lightgbm,
    author = {Ke, Guolin and Meng, Qi and Finley, Thomas and Wang, Taifeng and Chen, Wei and Ma, Weidong and Ye, Qiwei and Liu, Tie-Yan},
    booktitle = {Advances in Neural Information Processing Systems},
    editor = {I. Guyon and U. Von Luxburg and S. Bengio and H. Wallach and R. Fergus and S. Vishwanathan and R. Garnett},
    pages = {},
    publisher = {Curran Associates, Inc.},
    title = {LightGBM: A Highly Efficient Gradient Boosting Decision Tree},
    url = {https://proceedings.neurips.cc/paper_files/paper/2017/file/6449f44a102fde848669bdd9eb6b76fa-Paper.pdf},
    volume = {30},
    year = {2017}
}

@inproceedings{catboost,
 author = {Prokhorenkova, Liudmila and Gusev, Gleb and Vorobev, Aleksandr and Dorogush, Anna Veronika and Gulin, Andrey},
 booktitle = {Advances in Neural Information Processing Systems},
 editor = {S. Bengio and H. Wallach and H. Larochelle and K. Grauman and N. Cesa-Bianchi and R. Garnett},
 pages = {},
 publisher = {Curran Associates, Inc.},
 title = {CatBoost: unbiased boosting with categorical features},
 url = {https://proceedings.neurips.cc/paper_files/paper/2018/file/14491b756b3a51daac41c24863285549-Paper.pdf},
 volume = {31},
 year = {2018}
}

@misc{catboost_gp,
      title={Gradient Boosting Performs Gaussian Process Inference}, 
      author={Aleksei Ustimenko and Artem Beliakov and Liudmila Prokhorenkova},
      year={2023},
      eprint={2206.05608},
      archivePrefix={arXiv},
      primaryClass={cs.LG},
      url={https://arxiv.org/abs/2206.05608}, 
}

@InProceedings{ustimenko2020stochasticrank,
  title = 	 {StochasticRank: Global Optimization of Scale-Free Discrete Functions},
  author =       {Ustimenko, Aleksei and Prokhorenkova, Liudmila},
  booktitle = 	 {Proceedings of the 37th International Conference on Machine Learning},
  pages = 	 {9669--9679},
  year = 	 {2020},
  editor = 	 {III, Hal Daumé and Singh, Aarti},
  volume = 	 {119},
  series = 	 {Proceedings of Machine Learning Research},
  month = 	 {13--18 Jul},
  publisher =    {PMLR},
  url = 	 {https://proceedings.mlr.press/v119/ustimenko20a.html}
}

@article{TIAN2020150,
    title = {Credit Risk Assessment based on Gradient Boosting Decision Tree},
    journal = {Procedia Computer Science},
    volume = {174},
    pages = {150-160},
    year = {2020},
    note = {2019 International Conference on Identification, Information and Knowledge in the Internet of Things},
    issn = {1877-0509},
    doi = {https://doi.org/10.1016/j.procs.2020.06.070},
    url = {https://www.sciencedirect.com/science/article/pii/S1877050920315842},
    author = {Zhenya Tian and Jialiang Xiao and Haonan Feng and Yutian Wei}
}

@article{gbt_federated_healthcare,
    AUTHOR = {Wassan, Sobia and Suhail, Beenish and Mubeen, Riaqa and Raj, Bhavana and Agarwal, Ujjwal and Khatri, Eti and Gopinathan, Sujith and Dhiman, Gaurav},
    TITLE = {Gradient Boosting for Health IoT Federated Learning},
    JOURNAL = {Sustainability},
    VOLUME = {14},
    YEAR = {2022},
    NUMBER = {24},
    ARTICLE-NUMBER = {16842},
    URL = {https://www.mdpi.com/2071-1050/14/24/16842},
    ISSN = {2071-1050},
    DOI = {10.3390/su142416842}
}

@inproceedings{gbt_healthcare,
    author = {Ma, Handong and Cao, Jiahang and Fang, Yuchen and Zhang, Weinan and Sheng, Wenbo and Zhang, Shaodian and Yu, Yong},
    title = {Retrieval-Based Gradient Boosting Decision Trees for Disease Risk Assessment},
    year = {2022},
    isbn = {9781450393850},
    publisher = {Association for Computing Machinery},
    address = {New York, NY, USA},
    url = {https://doi.org/10.1145/3534678.3539052},
    doi = {10.1145/3534678.3539052},
    booktitle = {Proceedings of the 28th ACM SIGKDD Conference on Knowledge Discovery and Data Mining},
    pages = {3468–3476},
    numpages = {9},
    location = {Washington DC, USA},
    series = {KDD '22}
}

@Article{gbt_reliability_healthcare,
author={Seto, Hiroe
and Oyama, Asuka
and Kitora, Shuji
and Toki, Hiroshi
and Yamamoto, Ryohei
and Kotoku, Jun'ichi
and Haga, Akihiro
and Shinzawa, Maki
and Yamakawa, Miyae
and Fukui, Sakiko
and Moriyama, Toshiki},
title={Gradient boosting decision tree becomes more reliable than logistic regression in predicting probability for diabetes with big data},
journal={Scientific Reports},
year={2022},
month={Oct},
day={11},
volume={12},
number={1},
pages={15889},
issn={2045-2322},
doi={10.1038/s41598-022-20149-z},
url={https://doi.org/10.1038/s41598-022-20149-z}
}

@article{stable-baselines3,
    author  = {Antonin Raffin and Ashley Hill and Adam Gleave and Anssi Kanervisto and Maximilian Ernestus and Noah Dormann},
    title   = {Stable-Baselines3: Reliable Reinforcement Learning Implementations},
    journal = {Journal of Machine Learning Research},
    year    = {2021},
    volume  = {22},
    number  = {268},
    pages   = {1-8},
    url     = {http://jmlr.org/papers/v22/20-1364.html}
}

@inproceedings{MinigridMiniworld23,
 author = {Chevalier-Boisvert, Maxime and Dai, Bolun and Towers, Mark and Perez-Vicente, Rodrigo and Willems, Lucas and Lahlou, Salem and Pal, Suman and Castro, Pablo Samuel and Terry, J},
 booktitle = {Advances in Neural Information Processing Systems},
 editor = {A. Oh and T. Naumann and A. Globerson and K. Saenko and M. Hardt and S. Levine},
 pages = {73383--73394},
 publisher = {Curran Associates, Inc.},
 title = {Minigrid \& Miniworld: Modular \& Customizable Reinforcement Learning Environments for Goal-Oriented Tasks},
 url = {https://proceedings.neurips.cc/paper_files/paper/2023/file/e8916198466e8ef218a2185a491b49fa-Paper-Datasets_and_Benchmarks.pdf},
 volume = {36},
 year = {2023}
}

@misc{towers2024gymnasium,
      title={Gymnasium: A Standard Interface for Reinforcement Learning Environments}, 
      author={Mark Towers and Ariel Kwiatkowski and Jordan Terry and John U. Balis and Gianluca De Cola and Tristan Deleu and Manuel Goulão and Andreas Kallinteris and Markus Krimmel and Arjun KG and Rodrigo Perez-Vicente and Andrea Pierré and Sander Schulhoff and Jun Jet Tai and Hannah Tan and Omar G. Younis},
      year={2024},
      eprint={2407.17032},
      archivePrefix={arXiv},
      primaryClass={cs.LG},
      url={https://arxiv.org/abs/2407.17032}, 
}

@article{kurach2020google, title={Google Research Football: A Novel Reinforcement Learning Environment}, volume={34}, url={https://ojs.aaai.org/index.php/AAAI/article/view/5878}, DOI={10.1609/aaai.v34i04.5878}, abstractNote={&lt;p&gt;Recent progress in the field of reinforcement learning has been accelerated by virtual learning environments such as video games, where novel algorithms and ideas can be quickly tested in a safe and reproducible manner. We introduce the &lt;em&gt;Google Research Football Environment&lt;/em&gt;, a new reinforcement learning environment where agents are trained to play football in an advanced, physics-based 3D simulator. The resulting environment is challenging, easy to use and customize, and it is available under a permissive open-source license. In addition, it provides support for multiplayer and multi-agent experiments. We propose three full-game scenarios of varying difficulty with the &lt;em&gt;Football Benchmarks&lt;/em&gt; and report baseline results for three commonly used reinforcement algorithms (IMPALA, PPO, and Ape-X DQN). We also provide a diverse set of simpler scenarios with the &lt;em&gt;Football Academy&lt;/em&gt; and showcase several promising research directions.&lt;/p&gt;}, number={04}, journal={Proceedings of the AAAI Conference on Artificial Intelligence}, author={Kurach, Karol and Raichuk, Anton and Stanczyk, Piotr and Zajac, Michal and Bachem, Olivier and Espeholt, Lasse and Riquelme, Carlos and Vincent, Damien and Michalski, Marcin and Bousquet, Olivier and Gelly, Sylvain}, year={2020}, month={Apr.}, pages={4501-4510} }

@misc{abel2016exploratory,
      title={Exploratory Gradient Boosting for Reinforcement Learning in Complex Domains}, 
      author={David Abel and Alekh Agarwal and Fernando Diaz and Akshay Krishnamurthy and Robert E. Schapire},
      year={2016},
      eprint={1603.04119},
      archivePrefix={arXiv},
      primaryClass={cs.AI},
      url={https://arxiv.org/abs/1603.04119}, 
}

@inproceedings{brukhim2023boosting,
 author = {Brukhim, Nataly and Hazan, Elad and Singh, Karan},
 booktitle = {Advances in Neural Information Processing Systems},
 editor = {S. Koyejo and S. Mohamed and A. Agarwal and D. Belgrave and K. Cho and A. Oh},
 pages = {33806--33817},
 publisher = {Curran Associates, Inc.},
 title = {A Boosting Approach to Reinforcement Learning},
 url = {https://proceedings.neurips.cc/paper_files/paper/2022/file/daf8364f0715a41a469c677c0adc4754-Paper-Conference.pdf},
 volume = {35},
 year = {2022}
}

@inproceedings{pmlr-v206-london23a,
    title = {Boosted Off-Policy Learning},
    author = {London, Ben and Lu, Levi and Sandler, Ted and Joachims, Thorsten},
    booktitle = {Proceedings of The 26th International Conference on Artificial Intelligence and Statistics},
    pages = {5614--5640},
    year = {2023},
    editor = {Ruiz, Francisco and Dy, Jennifer and van de Meent, Jan-Willem},
    volume = {206},
    series = {Proceedings of Machine Learning Research},
    month = {25--27 Apr},
    publisher = {PMLR},
    url = {https://proceedings.mlr.press/v206/london23a.html}
}

@inproceedings{grinsztajn2022treebased,
 author = {Grinsztajn, Leo and Oyallon, Edouard and Varoquaux, Gael},
 booktitle = {Advances in Neural Information Processing Systems},
 editor = {S. Koyejo and S. Mohamed and A. Agarwal and D. Belgrave and K. Cho and A. Oh},
 pages = {507--520},
 publisher = {Curran Associates, Inc.},
 title = {Why do tree-based models still outperform deep learning on typical tabular data?},
 url = {https://proceedings.neurips.cc/paper_files/paper/2022/file/0378c7692da36807bdec87ab043cdadc-Paper-Datasets_and_Benchmarks.pdf},
 volume = {35},
 year = {2022}
}

@misc{rl-zoo3,
    author = {Raffin, Antonin},
    title = {RL Baselines3 Zoo},
    year = {2020},
    publisher = {GitHub},
    journal = {GitHub repository},
    howpublished = {\url{https://github.com/DLR-RM/rl-baselines3-zoo}}
}

@misc{ddpg,
      title={Continuous control with deep reinforcement learning}, 
      author={Timothy P. Lillicrap and Jonathan J. Hunt and Alexander Pritzel and Nicolas Heess and Tom Erez and Yuval Tassa and David Silver and Daan Wierstra},
      year={2019},
      eprint={1509.02971},
      archivePrefix={arXiv},
      primaryClass={cs.LG},
      url={https://arxiv.org/abs/1509.02971}, 
}

@book{Sutton1998,
    author = {Sutton, Richard S. and Barto, Andrew G.},
    edition = {Second},
    publisher = {The MIT Press},
    title = {Reinforcement Learning: An Introduction},
    url = {http://incompleteideas.net/book/the-book-2nd.html},
    year = {2018}
}

@inproceedings{katzir2021netdnf,
    title={Net-DNF: Effective Deep Modeling of Tabular Data},
    author={Liran Katzir and Gal Elidan and Ran El-Yaniv},
    booktitle={International Conference on Learning Representations},
    year={2021},
    url={https://openreview.net/forum?id=73WTGs96kho}
}

@inproceedings{gorishniy2023revisiting,
 author = {Gorishniy, Yury and Rubachev, Ivan and Khrulkov, Valentin and Babenko, Artem},
 booktitle = {Advances in Neural Information Processing Systems},
 editor = {M. Ranzato and A. Beygelzimer and Y. Dauphin and P.S. Liang and J. Wortman Vaughan},
 pages = {18932--18943},
 publisher = {Curran Associates, Inc.},
 title = {Revisiting Deep Learning Models for Tabular Data},
 url = {https://proceedings.neurips.cc/paper_files/paper/2021/file/9d86d83f925f2149e9edb0ac3b49229c-Paper.pdf},
 volume = {34},
 year = {2021}
}

@misc{malinin2021uncertainty,
      title={Uncertainty in Gradient Boosting via Ensembles}, 
      author={Andrey Malinin and Liudmila Prokhorenkova and Aleksei Ustimenko},
      year={2021},
      eprint={2006.10562},
      archivePrefix={arXiv},
      primaryClass={cs.LG},
      url={https://arxiv.org/abs/2006.10562}, 
}

@InProceedings{duan2020ngboost,
  title = 	 {{NGB}oost: Natural Gradient Boosting for Probabilistic Prediction},
  author =       {Duan, Tony and Anand, Avati and Ding, Daisy Yi and Thai, Khanh K. and Basu, Sanjay and Ng, Andrew and Schuler, Alejandro},
  booktitle = 	 {Proceedings of the 37th International Conference on Machine Learning},
  pages = 	 {2690--2700},
  year = 	 {2020},
  editor = 	 {III, Hal Daumé and Singh, Aarti},
  volume = 	 {119},
  series = 	 {Proceedings of Machine Learning Research},
  month = 	 {13--18 Jul},
  publisher =    {PMLR},
  url = 	 {https://proceedings.mlr.press/v119/duan20a.html}
}

@InProceedings{lyzhin2023tricks,
  title = 	 {Which Tricks are Important for Learning to Rank?},
  author =       {Lyzhin, Ivan and Ustimenko, Aleksei and Gulin, Andrey and Prokhorenkova, Liudmila},
  booktitle = 	 {Proceedings of the 40th International Conference on Machine Learning},
  pages = 	 {23264--23278},
  year = 	 {2023},
  editor = 	 {Krause, Andreas and Brunskill, Emma and Cho, Kyunghyun and Engelhardt, Barbara and Sabato, Sivan and Scarlett, Jonathan},
  volume = 	 {202},
  series = 	 {Proceedings of Machine Learning Research},
  month = 	 {23--29 Jul},
  publisher =    {PMLR},
  url = 	 {https://proceedings.mlr.press/v202/lyzhin23a.html}
}

@misc{ivanov2021boost,
      title={Boost then Convolve: Gradient Boosting Meets Graph Neural Networks}, 
      author={Sergei Ivanov and Liudmila Prokhorenkova},
      year={2021},
      eprint={2101.08543},
      archivePrefix={arXiv},
      primaryClass={cs.LG},
      url={https://arxiv.org/abs/2101.08543}, 
}

@article{sigrist2022gaussian,
  author  = {Fabio Sigrist},
  title   = {Gaussian Process Boosting},
  journal = {Journal of Machine Learning Research},
  year    = {2022},
  volume  = {23},
  number  = {232},
  pages   = {1--46},
  url     = {http://jmlr.org/papers/v23/20-322.html}
}

@article{Bellemare_2013,
    title={The Arcade Learning Environment: An Evaluation Platform for General Agents},
    volume={47},
    ISSN={1076-9757},
    url={http://dx.doi.org/10.1613/jair.3912},
    doi={10.1613/jair.3912},
    journal={Journal of Artificial Intelligence Research},
    publisher={AI Access Foundation},
    author={Bellemare, M. G. and Naddaf, Y. and Veness, J. and Bowling, M.},
    year={2013},
    month=jun,
    pages={253–279}
}

@misc{xgboost_demo,
    author = {Tianqi Chen},
    title = {Machine Learning Challenge Winning Solutions},
    year = {2023},
    note = {\url{https://github.com/dmlc/xgboost/tree/master/demo#machine-learning-challenge-winning-solutions}}
}

@inproceedings{NEURIPS2023_f06d5ebd,
 author = {McElfresh, Duncan and Khandagale, Sujay and Valverde, Jonathan and Prasad C, Vishak and Ramakrishnan, Ganesh and Goldblum, Micah and White, Colin},
 booktitle = {Advances in Neural Information Processing Systems},
 editor = {A. Oh and T. Naumann and A. Globerson and K. Saenko and M. Hardt and S. Levine},
 pages = {76336--76369},
 publisher = {Curran Associates, Inc.},
 title = {When Do Neural Nets Outperform Boosted Trees on Tabular Data?},
 url = {https://proceedings.neurips.cc/paper_files/paper/2023/file/f06d5ebd4ff40b40dd97e30cee632123-Paper-Datasets_and_Benchmarks.pdf},
 volume = {36},
 year = {2023}
}

@misc{somepalli2021saint,
      title={SAINT: Improved Neural Networks for Tabular Data via Row Attention and Contrastive Pre-Training}, 
      author={Gowthami Somepalli and Micah Goldblum and Avi Schwarzschild and C. Bayan Bruss and Tom Goldstein},
      year={2021},
      eprint={2106.01342},
      archivePrefix={arXiv},
      primaryClass={cs.LG},
      url={https://arxiv.org/abs/2106.01342}, 
}

@article{ota2021training,
author = {Ota, Kei and Jha, Devesh K. and Kanezaki, Asako},
title = {A framework for training larger networks for deep Reinforcement learning},
year = {2024},
issue_date = {Sep 2024},
publisher = {Kluwer Academic Publishers},
address = {USA},
volume = {113},
number = {9},
issn = {0885-6125},
url = {https://doi.org/10.1007/s10994-024-06547-6},
doi = {10.1007/s10994-024-06547-6},
journal = {Mach. Learn.},
month = jun,
pages = {6115–6139},
numpages = {25}
}

@article{tessler2017deep, title={A Deep Hierarchical Approach to Lifelong Learning in Minecraft}, volume={31}, url={https://ojs.aaai.org/index.php/AAAI/article/view/10744}, DOI={10.1609/aaai.v31i1.10744}, abstractNote={ &lt;p&gt; We propose a lifelong learning system that has the ability to reuse and transfer knowledge from one task to another while efficiently retaining the previously learned knowledge-base. Knowledge is transferred by learning reusable skills to solve tasks in Minecraft, a popular video game which is an unsolved and high-dimensional lifelong learning problem. These reusable skills, which we refer to as Deep Skill Networks, are then incorporated into our novel Hierarchical Deep Reinforcement Learning Network (H-DRLN) architecture using two techniques: (1) a deep skill array and (2) skill distillation, our novel variation of policy distillation (Rusu et. al. 2015) for learning skills. Skill distillation enables the H-DRLN to efficiently retain knowledge and therefore scale in lifelong learning, by accumulating knowledge and encapsulating multiple reusable skills into a single distilled network. The H-DRLN exhibits superior performance and lower learning sample complexity compared to the regular Deep Q Network (Mnih et. al. 2015) in sub-domains of Minecraft. &lt;/p&gt; }, number={1}, journal={Proceedings of the AAAI Conference on Artificial Intelligence}, author={Tessler, Chen and Givony, Shahar and Zahavy, Tom and Mankowitz, Daniel and Mannor, Shie}, year={2017}, month={Feb.} }

@inproceedings{jeffares2024deeplearningtelescopinglens,
 author = {Jeffares, Alan and Curth, Alicia and van der Schaar, Mihaela},
 booktitle = {Advances in Neural Information Processing Systems},
 editor = {A. Globerson and L. Mackey and D. Belgrave and A. Fan and U. Paquet and J. Tomczak and C. Zhang},
 pages = {123498--123533},
 publisher = {Curran Associates, Inc.},
 title = {Deep Learning Through A Telescoping Lens: A Simple Model Provides Empirical Insights On Grokking, Gradient Boosting \&amp; Beyond},
 url = {https://proceedings.neurips.cc/paper_files/paper/2024/file/df334022279996b07e0870a629c18857-Paper-Conference.pdf},
 volume = {37},
 year = {2024}
}

@misc{zabergja2024tabulardataattentionneed,
      title={Tabular Data: Is Attention All You Need?}, 
      author={Guri Zabërgja and Arlind Kadra and Josif Grabocka},
      year={2024},
      eprint={2402.03970},
      archivePrefix={arXiv},
      primaryClass={cs.LG},
      url={https://arxiv.org/abs/2402.03970}, 
}

@incollection{shap,
title = {A Unified Approach to Interpreting Model Predictions},
author = {Lundberg, Scott M and Lee, Su-In},
booktitle = {Advances in Neural Information Processing Systems 30},
editor = {I. Guyon and U. V. Luxburg and S. Bengio and H. Wallach and R. Fergus and S. Vishwanathan and R. Garnett},
pages = {4765--4774},
year = {2017},
publisher = {Curran Associates, Inc.},
url = {http://papers.nips.cc/paper/7062-a-unified-approach-to-interpreting-model-predictions.pdf}
}

@InProceedings{certifiedadversarialrobustnessdeep,
  title = 	 {Certified Adversarial Robustness for Deep Reinforcement Learning},
  author =       {L\"utjens, Bj\"orn and Everett, Michael and How, Jonathan P.},
  booktitle = 	 {Proceedings of the Conference on Robot Learning},
  pages = 	 {1328--1337},
  year = 	 {2020},
  editor = 	 {Kaelbling, Leslie Pack and Kragic, Danica and Sugiura, Komei},
  volume = 	 {100},
  series = 	 {Proceedings of Machine Learning Research},
  month = 	 {30 Oct--01 Nov},
  publisher =    {PMLR},
  url = 	 {https://proceedings.mlr.press/v100/lutjens20a.html}
}

@inproceedings{carlini2017adversarialexampleseasilydetected,
author = {Carlini, Nicholas and Wagner, David},
title = {Adversarial Examples Are Not Easily Detected: Bypassing Ten Detection Methods},
year = {2017},
isbn = {9781450352024},
publisher = {Association for Computing Machinery},
address = {New York, NY, USA},
url = {https://doi.org/10.1145/3128572.3140444},
doi = {10.1145/3128572.3140444},
booktitle = {Proceedings of the 10th ACM Workshop on Artificial Intelligence and Security},
pages = {3–14},
numpages = {12},
location = {Dallas, Texas, USA},
series = {AISec '17}
}

@inproceedings{poesia2021contrastivereinforcementlearningsymbolic,
 author = {Poesia, Gabriel and Dong, WenXin and Goodman, Noah},
 booktitle = {Advances in Neural Information Processing Systems},
 editor = {M. Ranzato and A. Beygelzimer and Y. Dauphin and P.S. Liang and J. Wortman Vaughan},
 pages = {15946--15956},
 publisher = {Curran Associates, Inc.},
 title = {Contrastive Reinforcement Learning of Symbolic Reasoning Domains},
 url = {https://proceedings.neurips.cc/paper_files/paper/2021/file/859555c74e9afd45ab771c615c1e49a6-Paper.pdf},
 volume = {34},
 year = {2021}
}

@misc{xu2021neuralnetworksextrapolatefeedforward,
      title={How Neural Networks Extrapolate: From Feedforward to Graph Neural Networks}, 
      author={Keyulu Xu and Mozhi Zhang and Jingling Li and Simon S. Du and Ken-ichi Kawarabayashi and Stefanie Jegelka},
      year={2021},
      eprint={2009.11848},
      archivePrefix={arXiv},
      primaryClass={cs.LG},
      url={https://arxiv.org/abs/2009.11848}, 
}

@misc{meinke2020neuralnetworksprovablyknow,
      title={Towards neural networks that provably know when they don't know}, 
      author={Alexander Meinke and Matthias Hein},
      year={2020},
      eprint={1909.12180},
      archivePrefix={arXiv},
      primaryClass={cs.LG},
      url={https://arxiv.org/abs/1909.12180}, 
}

@InProceedings{ulmer2021knowlimitsuncertaintyestimation,
  title = 	 {Know your limits: Uncertainty estimation with ReLU classifiers fails at reliable OOD detection},
  author =       {Ulmer, Dennis and Cin\`a, Giovanni},
  booktitle = 	 {Proceedings of the Thirty-Seventh Conference on Uncertainty in Artificial Intelligence},
  pages = 	 {1766--1776},
  year = 	 {2021},
  editor = 	 {de Campos, Cassio and Maathuis, Marloes H.},
  volume = 	 {161},
  series = 	 {Proceedings of Machine Learning Research},
  month = 	 {27--30 Jul},
  publisher =    {PMLR},
  url = 	 {https://proceedings.mlr.press/v161/ulmer21a.html}
}

@inproceedings{kadra2021welltunedsimplenetsexcel,
 author = {Kadra, Arlind and Lindauer, Marius and Hutter, Frank and Grabocka, Josif},
 booktitle = {Advances in Neural Information Processing Systems},
 editor = {M. Ranzato and A. Beygelzimer and Y. Dauphin and P.S. Liang and J. Wortman Vaughan},
 pages = {23928--23941},
 publisher = {Curran Associates, Inc.},
 title = {Well-tuned Simple Nets Excel on Tabular Datasets},
 url = {https://proceedings.neurips.cc/paper_files/paper/2021/file/c902b497eb972281fb5b4e206db38ee6-Paper.pdf},
 volume = {34},
 year = {2021}
}

@article{arik2020tabnetattentiveinterpretabletabular, title={TabNet: Attentive Interpretable Tabular Learning}, volume={35}, url={https://ojs.aaai.org/index.php/AAAI/article/view/16826}, DOI={10.1609/aaai.v35i8.16826}, abstractNote={We propose a novel high-performance and interpretable canonical deep tabular data learning architecture, TabNet. TabNet uses sequential attention to choose which features to reason from at each decision step, enabling interpretability and more efficient learning as the learning capacity is used for the most salient features. We demonstrate that TabNet outperforms other variants on a wide range of non-performance-saturated tabular datasets and yields interpretable feature attributions plus insights into its global behavior. Finally, we demonstrate self-supervised learning for tabular data, significantly improving performance when unlabeled data is abundant.}, number={8}, journal={Proceedings of the AAAI Conference on Artificial Intelligence}, author={Arik, Sercan O. and Pfister, Tomas}, year={2021}, month={May}, pages={6679-6687} }

@misc{hollmann2023tabpfntransformersolvessmall,
      title={TabPFN: A Transformer That Solves Small Tabular Classification Problems in a Second}, 
      author={Noah Hollmann and Samuel Müller and Katharina Eggensperger and Frank Hutter},
      year={2023},
      eprint={2207.01848},
      archivePrefix={arXiv},
      primaryClass={cs.LG},
      url={https://arxiv.org/abs/2207.01848}, 
}

@article{Borisov_2024,
  title={Deep Neural Networks and Tabular Data: A Survey},
  author={Vadim Borisov and Tobias Leemann and Kathrin Se{\ss}ler and Johannes Haug and Martin Pawelczyk and Gjergji Kasneci},
  journal={IEEE Transactions on Neural Networks and Learning Systems},
  year={2021},
  volume={35},
  pages={7499-7519},
  url={https://api.semanticscholar.org/CorpusID:238353897}
}

@article{shwartzziv2021tabulardatadeeplearning,
title = {Tabular data: Deep learning is not all you need},
journal = {Information Fusion},
volume = {81},
pages = {84-90},
year = {2022},
issn = {1566-2535},
doi = {https://doi.org/10.1016/j.inffus.2021.11.011},
url = {https://www.sciencedirect.com/science/article/pii/S1566253521002360},
author = {Ravid Shwartz-Ziv and Amitai Armon}
}

@inproceedings{Beyazit2023AnIB,
 author = {Beyazit, Ege and Kozaczuk, Jonathan and Li, Bo and Wallace, Vanessa and Fadlallah, Bilal},
 booktitle = {Advances in Neural Information Processing Systems},
 editor = {A. Oh and T. Naumann and A. Globerson and K. Saenko and M. Hardt and S. Levine},
 pages = {43108--43135},
 publisher = {Curran Associates, Inc.},
 title = {An Inductive Bias for Tabular Deep Learning},
 url = {https://proceedings.neurips.cc/paper_files/paper/2023/file/8671b6dffc08b4fcf5b8ce26799b2bef-Paper-Conference.pdf},
 volume = {36},
 year = {2023}
}

@article{henderson2019deepreinforcementlearningmatters, title={Deep Reinforcement Learning That Matters}, volume={32}, url={https://ojs.aaai.org/index.php/AAAI/article/view/11694}, DOI={10.1609/aaai.v32i1.11694}, abstractNote={ &lt;p&gt; In recent years, significant progress has been made in solving challenging problems across various domains using deep reinforcement learning (RL). Reproducing existing work and accurately judging the improvements offered by novel methods is vital to sustaining this progress. Unfortunately, reproducing results for state-of-the-art deep RL methods is seldom straightforward. In particular, non-determinism in standard benchmark environments, combined with variance intrinsic to the methods, can make reported results tough to interpret. Without significance metrics and tighter standardization of experimental reporting, it is difficult to determine whether improvements over the prior state-of-the-art are meaningful. In this paper, we investigate challenges posed by reproducibility, proper experimental techniques, and reporting procedures. We illustrate the variability in reported metrics and results when comparing against common baselines and suggest guidelines to make future results in deep RL more reproducible. We aim to spur discussion about how to ensure continued progress in the field by minimizing wasted effort stemming from results that are non-reproducible and easily misinterpreted. &lt;/p&gt; }, number={1}, journal={Proceedings of the AAAI Conference on Artificial Intelligence}, author={Henderson, Peter and Islam, Riashat and Bachman, Philip and Pineau, Joelle and Precup, Doina and Meger, David}, year={2018}, month={Apr.} }

@misc{andrychowicz2020mattersonpolicyreinforcementlearning,
      title={What Matters In On-Policy Reinforcement Learning? A Large-Scale Empirical Study}, 
      author={Marcin Andrychowicz and Anton Raichuk and Piotr Stanczyk and Manu Orsini and Sertan Girgin and Raphael Marinier and Léonard Hussenot and Matthieu Geist and Olivier Pietquin and Marcin Michalski and Sylvain Gelly and Olivier Bachem},
      year={2020},
      eprint={2006.05990},
      archivePrefix={arXiv},
      primaryClass={cs.LG},
      url={https://arxiv.org/abs/2006.05990}, 
}

@Article{Hancock2020,
author={Hancock, John T.
and Khoshgoftaar, Taghi M.},
title={Survey on categorical data for neural networks},
journal={Journal of Big Data},
year={2020},
month={Apr},
day={10},
volume={7},
number={1},
pages={28},
issn={2196-1115},
doi={10.1186/s40537-020-00305-w},
url={https://doi.org/10.1186/s40537-020-00305-w}
}

@inproceedings{gorishniy2023embeddingsnumericalfeaturestabular,
 author = {Gorishniy, Yury and Rubachev, Ivan and Babenko, Artem},
 booktitle = {Advances in Neural Information Processing Systems},
 editor = {S. Koyejo and S. Mohamed and A. Agarwal and D. Belgrave and K. Cho and A. Oh},
 pages = {24991--25004},
 publisher = {Curran Associates, Inc.},
 title = {On Embeddings for Numerical Features in Tabular Deep Learning},
 url = {https://proceedings.neurips.cc/paper_files/paper/2022/file/9e9f0ffc3d836836ca96cbf8fe14b105-Paper-Conference.pdf},
 volume = {35},
 year = {2022}
}

@misc{paszke2019pytorchimperativestylehighperformance,
      title={PyTorch: An Imperative Style, High-Performance Deep Learning Library}, 
      author={Adam Paszke and Sam Gross and Francisco Massa and Adam Lerer and James Bradbury and Gregory Chanan and Trevor Killeen and Zeming Lin and Natalia Gimelshein and Luca Antiga and Alban Desmaison and Andreas Köpf and Edward Yang and Zach DeVito and Martin Raison and Alykhan Tejani and Sasank Chilamkurthy and Benoit Steiner and Lu Fang and Junjie Bai and Soumith Chintala},
      year={2019},
      eprint={1912.01703},
      archivePrefix={arXiv},
      primaryClass={cs.LG},
      url={https://arxiv.org/abs/1912.01703}, 
}

@book{shapley:book1952,
 ISBN = {9780691079356},
 URL = {http://www.jstor.org/stable/j.ctt1b9x1zv},
 author = {K. J. ARROW and E. W. BARANKIN and D. BLACKWELL and R. BOTT and N. DALKEY and M. DRESHER and D. GALE and D. B. GILLIES and I. GLICKSBERG and O. GROSS and S. KARLIN and H. W. KUHN and J. P. MAYBERRY and J. W. MILNOR and T. S. MOTZKIN and J. VON NEUMANN and H. RAIFFA and L. S. SHAPLEY and M. SHIFFMAN and F. M. STEWART and G. L. THOMPSON and R. M. THRALL},
 publisher = {Princeton University Press},
 title = {Contributions to the Theory of Games (AM-28), Volume II},
 urldate = {2025-04-15},
 year = {1953}
}

@inproceedings{DBLP:journals/corr/SchulmanMLJA15,
  author       = {John Schulman and
                  Philipp Moritz and
                  Sergey Levine and
                  Michael I. Jordan and
                  Pieter Abbeel},
  editor       = {Yoshua Bengio and
                  Yann LeCun},
  title        = {High-Dimensional Continuous Control Using Generalized Advantage Estimation},
  booktitle    = {4th International Conference on Learning Representations, {ICLR} 2016,
                  San Juan, Puerto Rico, May 2-4, 2016, Conference Track Proceedings},
  year         = {2016},
  url          = {http://arxiv.org/abs/1506.02438},
  bibsource    = {dblp computer science bibliography, https://dblp.org}
}
\bibliographystyle{icml2025}

\newpage
\appendix
\onecolumn
\section*{Appendix}
This appendix provides supplementary materials that support the findings and methodologies discussed in the main text. It is organized into five sections to present the full experiment results, implementation details, hyperparameters used during the experiments, training progression plots, and experimental plots, respectively. These materials offer detailed insights into the research process and outcomes, facilitating a deeper understanding and replication of the study.

\section{Multi-Objective Training With GBT}
In GBRL, we compute gradients for both the actor and critic objectives per timestep. We then concatenate these gradients: \(g_t := [g_{t,\text{actor}}, g_{t,\text{critic}}] \) and use them to build a decision tree. At each candidate node split, we evaluate a score function to decide the best split. However, since the actor and critic gradients can differ in magnitude, this score can become biased toward one objective. To address this, we explored two strategies:
\begin{itemize}
    \item Gradient normalization per output dimension with L2-based split scoring.
    \item Cosine similarity, which emphasizes gradient direction over magnitude.
\end{itemize}

\begin{figure*}[hbt!]
\centering
    \includegraphics[width=0.7\linewidth]{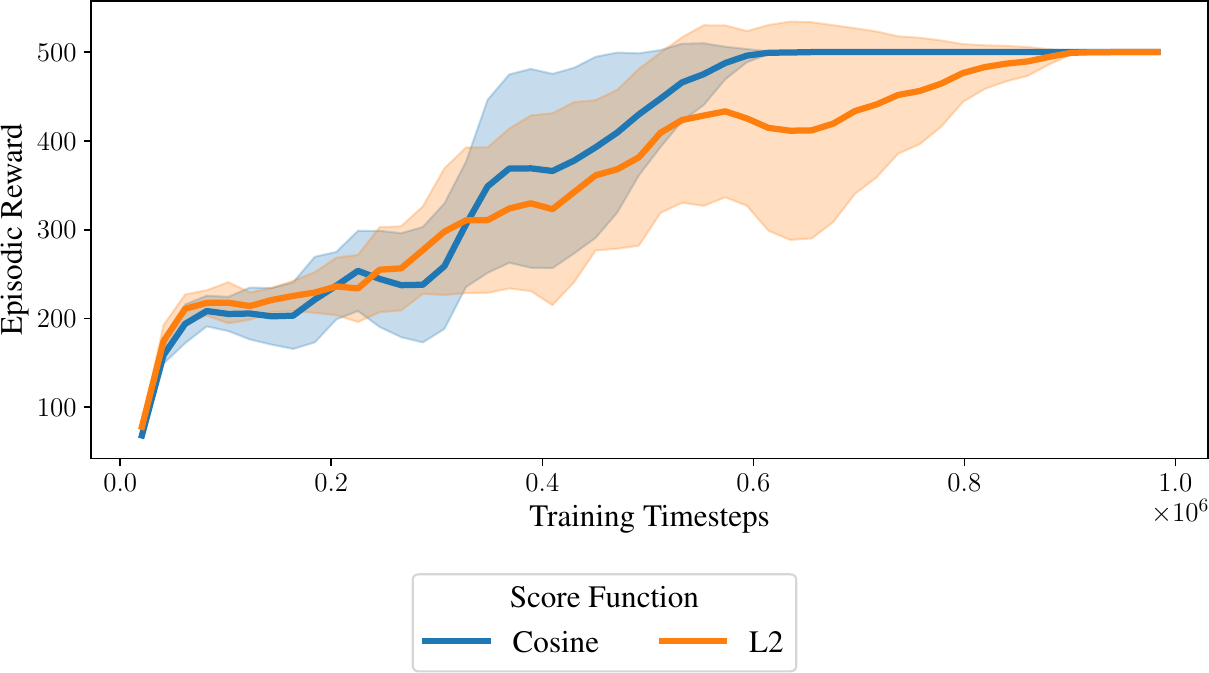}
\caption{\textbf{Multi-objective gradient split strategies in PPO-GBRL on CartPole-v1}. We compare gradient normalization and cosine similarity as methods for handling actor-critic gradient imbalance when building trees. Both strategies yield good performance, with cosine similarity achieving slightly faster convergence. Curves show average episodic reward over 1M timesteps.}
    \label{fig:ablation_score}
\end{figure*}

Both approaches are supported in our codebase and produce good results, while the cosine similarity performed slightly better (\cref{fig:ablation_score}). 

\section{Why Do Certain Algorithms Perform Differently Than Others?}
\label{sec:appendix:algo_diff}
\begin{figure*}[hbt!]
\centering
    \includegraphics[width=0.65\linewidth]{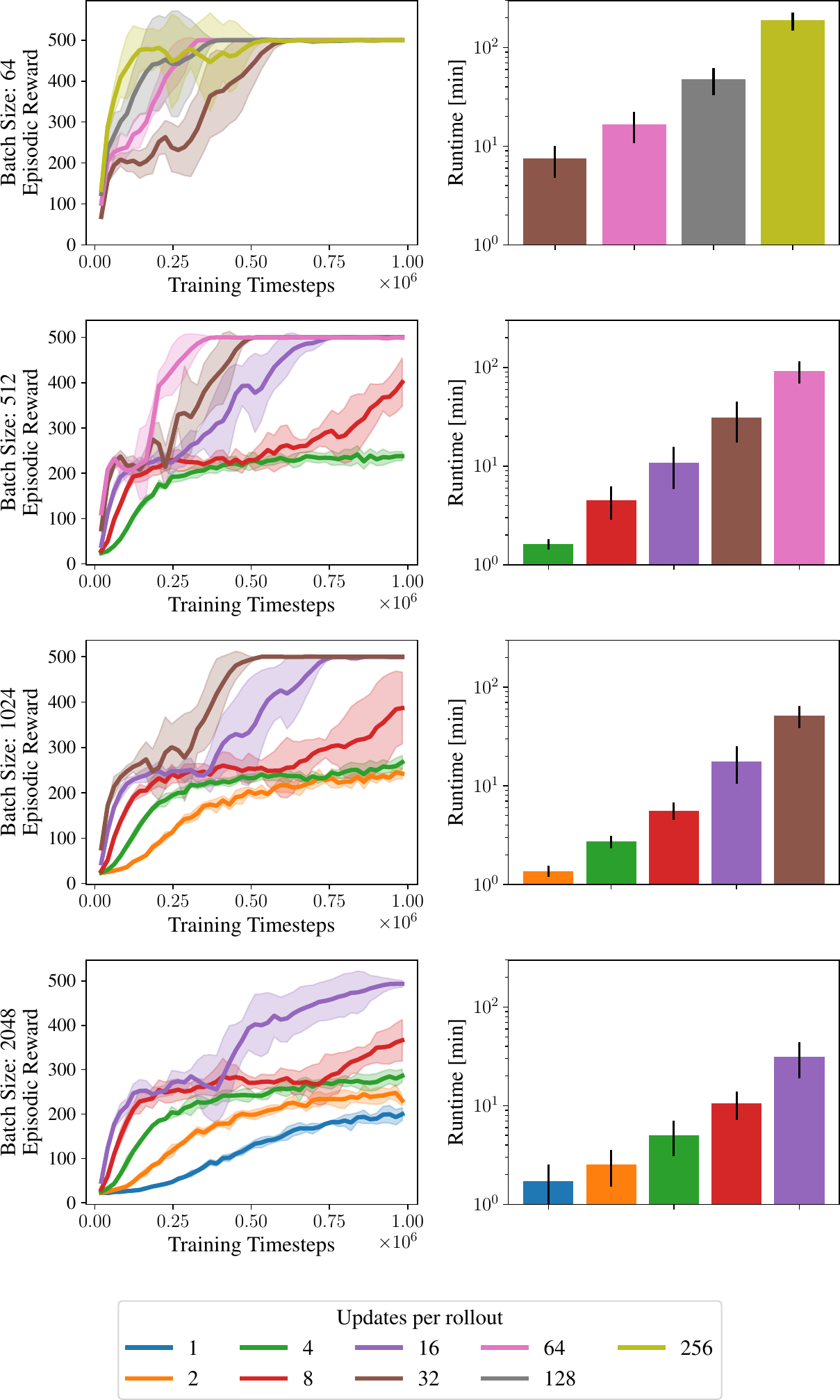}
\caption{\textbf{Effect of number of gradient updates per rollout on PPO-GBRL's performance in CartPole-v1}. We fix the batch size and the number of samples collected per rollout, while varying the number of PPO epochs, thereby changing the number of gradient updates per rollout. This improves sample efficiency, with more epochs leading to faster convergence. Results are shown over 1M timesteps. The number of samples collected per rollout was fixed to 2048 for all experiments.
}
    \label{fig:ablation_epochs}
\end{figure*}

Apart from inherent algorithmic differences, the main difference for GBRL is the number of fitted trees per update. This is analogous to the number of update steps in a neural network, since each tree represents a functional gradient step for the actor or critic. To illustrate this effect, we trained agents in the CartPole-v1 environment for 1M timesteps and varied the number of epochs while keeping the batch size and the rollout length fixed, thereby varying the number of updates per rollout. We repeated this experiment for several batch sizes while keeping the rollout length fixed at 2048. As shown in \cref{fig:ablation_epochs}, increasing the number of trees added per rollout improves performance, but increases computational cost. Moreover, a larger batch size also increases computational time, due to a larger number of samples used to build each tree. Hence, there is a clear trade-off between computational cost and performance. 

This trade-off also explains the difference in performance of various algorithms when combined with GBRL, as their update schemes inherently affect how many trees are fitted per rollout.

\begin{itemize}
    \item \textbf{A2C} - performs a single gradient step per rollout using the entire batch. In GBRL, this results in a single constructed tree per rollout. Since GBTs update the value and policy functions directly, such large, coarse updates can destabilize learning.
    \item \textbf{PPO} - in contrast, uses minibatches and multiple epochs per rollout to better control the trade-off. This allows GBRL to fit several small trees per rollout, yielding more stable and incremental updates.
    \item \textbf{AWR} - is an off-policy algorithm. Therefore, GBRL could potentially build many trees without adding sample complexity. However, we noticed that our best performing models had a huge computational cost to finish training. As a result, we resorted to reducing the number of gradient steps per update at the expense of performance. Future work could investigate methods for optimizing GBRL with a fixed tree budget (such as pruning and distillation), which may enable the applicability of more demanding RL algorithms, such as AWR, using GBRL.
\end{itemize}
In general, as each tree approximates a gradient step, GBRL benefits from frequent updates to refine its ensemble gradually. This makes it particularly well-suited to algorithms that support minibatching and multiple epochs per rollout.

\section{Implementation Details and Hyperparameters}
\label{sec:appendix:implementation_details}
This section provides implementation details, information regarding compute resources, and tables containing the hyperparameters used in our experiments enabling the reproducibility of our results. \cref{tab:hyperparams} lists GBRL hyperparameters for all experiments.

\subsection{Algorithm Details}
\label{appendix:algo_details}
Our GBRL method is implemented within the Stable Baselines3 framework \cite{stable-baselines3}, a widely used library for standardized reinforcement learning algorithms. For comparisons, we use baseline NNs from RL Zoo3 \cite{rl-zoo3}, which provides tuned hyperparameters per environment. For all experiments, we used a standard MLP with two hidden layers, each containing 64 units and tanh activations. The hidden layers are shared between the actor and the critic, and each component has an additional separate linear layer. This architecture is the default in Stable Baselines3 for the environments we tested and is widely used as a baseline in RL literature. We maintained this consistent architecture across all environments to ensure fair comparison.

\textbf{A2C} \cite{a2c} is a synchronous, on-policy Actor-Critic algorithm designed to improve learning stability.  The critic learns a value function, \(V(\state)\), used to estimate the advantage. This advantage is incorporated into the policy gradient updates, reducing variance and leading to smoother learning. The policy is updated using the following gradient: \(\nabla_\theta J(\pi_\theta) = \E[\nabla_\theta\log\pi_\theta(\action|\state) A(\state, \action)]\).


\textbf{AWR} \cite{awr} is an off-policy Actor-Critic algorithm. Provided a dataset \(\mathcal{D}\), AWR updates both the policy and the value through supervised learning. This dataset can be pre-defined and fixed (offline), or continually updated using the agent's experience (replay buffer). At each training iteration \(k\), AWR solves the following two regression problems:
\begin{equation*}
\begin{aligned}
    V_k &= \argmin_V \E_{\state, \action \sim \mathcal{D}}[\| G(\state, \action) - V(\state)\|_2^2] \,,\, \\
    \pi_{k+1} &= \argmax_{\pi} \E_{\state, \action \sim \mathcal{D}}[\log\pi(\action | \state)\exp(\frac{1}{\beta}A_k(\state, \action))] \,,
\end{aligned}
\end{equation*}
where \(G(\state, \action)\) represents the monte-carlo estimate or TD(\(\lambda\)) of the expected return \cite{Sutton1998}. 

\textbf{Advantage computation}: We use Generalized Advantage Estimation (GAE) as proposed by \citet{DBLP:journals/corr/SchulmanMLJA15} to compute the advantage: \(A_t = \sum_{l=0}^\infty (\gamma \lambda)^l \delta_{t+l}\) , where \(\delta_t = r_t + \gamma V(\state_{t+1}) - V(\state_t)\). The GAE parameters and for each environment are provided in \cref{tab:hyperparams} of the appendix.
Critic: We use on-policy actor-critic algorithms, where the critic predicts the value function (not action-value Q). We use a standard L2 loss between the critic's prediction and the target return estimate: \(L_{\text{critic}}(\theta) = ||V_{\theta}(\state_t) - G_t||^2\), where \(G_t\) is either a Monte Carlo return estimate or a bootstrapped n-step return, depending on the specific algorithm (PPO, A2C, or AWR).

\subsection{Environments}
\label{appendix:env_details}

\paragraph{Classic-Control \& Box2D.} We evaluate GBRL against NNs on all available classic-control tasks provided in Gymnasium \cite{towers2024gymnasium} and on LunarLander-v2. This includes three discrete action space environments: Acrobot-v1, CartPole-v1, MountainCar-v0, and two continuous action space environments: Pendulum-v1 and MountainCarContinuous-v0. We trained on both environments for 1M steps (1.5M for
LunarLander-v2). For NNs we used the hyperparameters provided in RL Baselines3 Zoo \cite{rl-zoo3}, except for AWR where we followed the hyperparameters described in the paper.   

\paragraph{Google Research Football \& Atari ram.} The Football domain \cite{kurach2020google} consists of a vectorized 115-dimensional observation space that summarizes the main aspects of the game and 19 discrete actions. We focus on its academy scenarios, which present situational tasks involving scoring a single goal. A standard reward of +1 is granted for scoring, and we employed the "Checkpoints" shaped reward structure. This structure provides additional points as the agent moves closer towards the goal, with a maximum reward of 2 per scenario.  The Atari-ram domain consists of a vectorized 128-dimensional observational space representing the 128-byte RAM state and up to 18 discrete actions. We trained agents in both domains for 10M timesteps.

\paragraph{MiniGrid.} The MiniGrid environment \cite{MinigridMiniworld23} is a 2D grid world with goal-oriented tasks requiring object interaction. The observation space consists of a 7x7 image representing the grid, a mission string, and the agent's direction. Each tile in the observed image contains a 3D tuple dictating an object's color, type, and state. All MiniGrid tasks emit a reward of +1 for successful completion and 0 otherwise. We trained our NN-based agents on a flattened observation space using the built-in one-hot wrapper, as specified by RL Baselines3 Zoo. For GBRL agents, we generated a 51-dimensional categorical observational space by encoding each unique tile tuple as a categorical string to represent the observed image. Categorical features were added for the agent's direction (up, left, right, down) and missions. All agents were trained for 1M timesteps, except for PutNear, FourRooms, and Fetch tasks, which were trained for 10M based on the reported values for PPO NN in RL Baselines3 Zoo. 

\paragraph{Variable Isolation Environment.} The goal of the variable isolation task is to separate a target variable within a linear equation. In the basic task variation, the initial state is a linear equation of the form \(ax + b = c\), where $a, b, c$ are digits within the range \([1-9]\) that can be either positive or negative. The action space is multi-discrete, where the agent chooses an action type, the digit to use, and the sign of the digit. The action types are: addition, subtraction, multiplication, and division. The agent receives a reward proportional to the number of steps taken in the episode if it successfully isolates \(x\), otherwise the reward is \(0\). 

In the two-variable linear equation variation, the initial state is a linear equation of the form \(ax + by + c = d\), and the goal is to isolate \(y\) on the left side. The actions are equivalent to the basic task, with the addition or subtraction of a digit multiplied by \(x\). Finally, in the balanced two-variable linear equation variation, the initial state is a linear equation of the form \(ax + b = cy + d\), and the goal is to isolate \(y\) on the left side. The actions are equivalent to the two-variable linear equation variation with the addition of the operations involving \(y\).  In all the environment variations, the agent observes the coefficients as inputs. For the two variable variations, we add a small reward bonus if the agent successfully moves a variable or a constant to the correct side. We trained GBRL for 15M steps and NNs for 30M steps.

\begin{table}[h!]
\centering
\begin{subtable}{\textwidth}
\resizebox{\textwidth}{!}{%
\begin{tabular}{ccccccccccc}
\toprule
 & batch size & clip range & ent coef & gae lambda & gamma & \begin{tabular}[c]{@{}c@{}}num epochs\end{tabular} & \begin{tabular}[c]{@{}c@{}}num steps\end{tabular} & \begin{tabular}[c]{@{}c@{}}num envs\end{tabular} & \begin{tabular}[c]{@{}c@{}}policy lr\end{tabular} & \begin{tabular}[c]{@{}c@{}}value lr\end{tabular} \\ 
\midrule
Acrobot & 512 & 0.2 & 0.0 & 0.94 & 0.99 & 20 & 128 & 16 & 0.16 & 0.034 \\
CartPole & 64 & 0.2 & 0.0 & 0.8 & 0.98 & 1 & 128 & 8 & 0.029 & 0.015 \\
LunarLander & 256 & 0.2 & 0.0033 & 0.98 & 0.999 & 20 & 512 & 16 & 0.031 & 0.003 \\
MountainCar & 256 & 0.2 & 0.033 & 0.98 & 0.999 & 20 & 512 & 16 & 0.031 & 0.003 \\
\begin{tabular}[c]{@{}c@{}}MountainCar Continuous\end{tabular} & 256 & 0.2 & 0.033 & 0.98 & 0.999 & 20 & 512 & 16 & 0.031 & 0.003 \\
Pendulum & 512 & 0.2 & 0.0 & 0.93 & 0.91 & 20 & 256 & 16 & 0.031 & 0.013 \\
Football & 512 & 0.2 & 0.0 & 0.95 & 0.998 & 10 & 256 & 16 & 0.033 & 0.006 \\
Atari-Ram & 64 & 0.92 & 8e-5 & 0.95 & 0.99 & 4 & 512 & 16 & 0.05 & 0.002 \\
MiniGrid & 512 & 0.2 & 0.0 & 0.95 & 0.99 & 20 & 256 & 16 & 0.17 & 0.01 \\
Variable Isolation & 128 & 0.62 & 0.08 & 0.95 & 0.99 & 20 & 256 & 16 & 0.18 & 0.006 \\
\bottomrule
\end{tabular}%
}
\caption{PPO. For continuous action spaces, we used log std init = -2 and log std lr = lin\_0.0017. We utilized gradient norm clipping for Gym environments. Specifically, 10 for the value gradients and 150 for the policy gradients.}
\label{tab:ppo_gbrl_hyperparams}
\end{subtable}

\vspace{15pt}

\begin{subtable}{\textwidth}
\centering
\resizebox{\textwidth}{!}{%
\begin{tabular}{cccccccccc}
\toprule
 & ent coef & gae lambda & gamma & \begin{tabular}[c]{@{}c@{}}num steps\end{tabular} & \begin{tabular}[c]{@{}c@{}}num envs\end{tabular} & \begin{tabular}[c]{@{}c@{}}policy lr\end{tabular} & \begin{tabular}[c]{@{}c@{}}value lr\end{tabular} & \begin{tabular}[c]{@{}c@{}}log std init\end{tabular} & \begin{tabular}[c]{@{}c@{}}log std lr\end{tabular} \\ 
\midrule
Acrobot & 0.0 & 1 & 0.99 & 8 & 4 & 0.79 & 0.031 & - & - \\
CartPole & 0.0 & 1 & 0.99 & 8 & 16 & 0.13 & 0.047 & - & - \\
LunarLander & 0.0 & 1 & 0.995 & 5 & 32 & 0.16 & 0.04 & - & - \\
MountainCar & 0.0 & 1 & 0.99 & 8 & 16 & 0.64 & 0.032 & - & - \\
\begin{tabular}[c]{@{}c@{}}MountainCar Continuous\end{tabular} & 0.0 & 1 & 0.995 & 128 & 16 & 0.0008 & 2.8e-6 & 0 & 0.0004 \\
Pendulum & 0.0 & 0.9 & 0.9 & 10 & 32 & 0.003 & 0.056 & -2 & 0.00018 \\
Football & 0.0004 & 0.95 & 0.998 & 128 & 8 & 0.87 & 0.017 & - & - \\
Atari-Ram & 0.0009 & 0.95 & 0.993 & 128 & 8 & 0.17 & 0.013 & - & - \\
MiniGrid & 0.0 & 0.95 & 0.99 & 10 & 128 & 0.34 & 0.039 & - & - \\
\bottomrule
\end{tabular}%
}
\caption{A2C}
\label{tab:a2c_gbrl_hyperparams}
\end{subtable}

\vspace{15pt}

\begin{subtable}{\textwidth}
\centering
\resizebox{\textwidth}{!}{%
\begin{tabular}{cccccccccccc}
\toprule
 & \begin{tabular}[c]{@{}c@{}}batch size\end{tabular} & ent coef & gae lambda & gamma & \begin{tabular}[c]{@{}c@{}}train freq\end{tabular} & \begin{tabular}[c]{@{}c@{}}gradient steps\end{tabular} & \begin{tabular}[c]{@{}c@{}}num envs\end{tabular} & \begin{tabular}[c]{@{}c@{}}policy lr\end{tabular} & \begin{tabular}[c]{@{}c@{}}value lr\end{tabular} & \begin{tabular}[c]{@{}c@{}}log std init\end{tabular} & \begin{tabular}[c]{@{}c@{}}log std lr\end{tabular} \\ 
\midrule
Acrobot & 1024 & 0.0 & 0.95 & 0.99 & 2000 & 150 & 1 & 0.05 & 0.1 & - & - \\
CartPole & 1024 & 0.0 & 0.95 & 0.99 & 2000 & 150 & 1 & 0.05 & 0.1 & - & - \\
LunarLander & 1024 & 0.0 & 0.95 & 0.99 & 2000 & 150 & 1 & 0.05 & 0.1 & - & - \\
MountainCar & 64 & 0.0 & 0.95 & 0.99 & 2000 & 150 & 1 & 0.64 & 0.032 & - & - \\
\begin{tabular}[c]{@{}c@{}}MountainCar Continuous\end{tabular} & 64 & 0.0 & 0.95 & 0.99 & 2000 & 150 & 1 & 0.089 & 0.083 & -2 & lin\_0.0017 \\
Pendulum & 1024 & 0.0 & 0.9 & 0.9 & 1000 & 50 & 1 & 0.003 & 0.07 & -2 & 0.0005 \\
Football & 512 & 0.03 & 0.95 & 0.99 & 750 & 10 & 1 & 0.09 & 0.00048 & - & - \\
Atari-Ram & 1024 & 0.0 & 0.95 & 0.993 & 2000 & 50 & 1 & 0.0779 & 0.0048 & - & - \\
MiniGrid & 1024 & 0.0 & 0.95 & 0.99 & 1500 & 25/100* & 1 & 0.0075 & 0.005 & - & - \\
\bottomrule
\end{tabular}%
}
\caption{AWR. For all envs, buffer size = 50,000,  $\beta$ = 0.05. *MiniGrid environments used 100 gradient steps for tasks trained for 1M steps, and 25 gradient steps for tasks trained for 10M steps, for a reduced tree size.}
\label{tab:awr_gbrl_hyperparams}
\end{subtable}
\caption{\textbf{GBRL hyperparameters} - NN represented by an MLP with two hidden layers.}
\label{tab:hyperparams}
\end{table}
\subsection{Compute Resources}
 All experiments were done on the NVIDIA NGC platform on a single NVIDIA V100-32GB GPU per experiment. Training time and compute requirements vary between algorithms and according to hyperparameters. The number of boosting iterations has the largest impact on both runtime and memory. GBRL experimental runs required from 1GB to 24GB of GPU memory. Moreover, runtime varied from 20 minutes for 1M timesteps training on classic environments and up to 5 days for 10M timesteps on Atari-ram. NN experimental runs required up to 3GB of GPU memory and runtime ranged from 10 minutes and up to 3 days. The total compute time for all experiments combined was approximately 1800 GPU hours. Additionally, the research project involved preliminary experiments and hyperparameter tuning, which required an estimated additional 168 GPU hours.

\section{Detailed Result Tables}
\label{appendix:results}
This section contains tables presenting the mean and standard deviation of the average episode reward for the final 100 episodes within each experiment. More specifically, \cref{tab:gym} presents results for Continuous Control \& Block2D environments, \cref{tab:football,tab:atari} present results for the high-dimensional vectorized environments, and \cref{tab:minigrid}
 presents results for the categorical environments.

\begin{table}[h!]
\centering
\caption{Continuous-Control and Box2D environments: Reported values are the mean and standard deviation of the average episode reward over the final 100 training episodes, aggregated across 5 random seeds.}
\resizebox{\textwidth}{!}{%
\begin{tabular}{
>{\columncolor[HTML]{FFFFFF}}c
>{\columncolor[HTML]{FFFFFF}}c
>{\columncolor[HTML]{FFFFFF}}c
>{\columncolor[HTML]{FFFFFF}}c 
>{\columncolor[HTML]{FFFFFF}}c 
>{\columncolor[HTML]{FFFFFF}}c 
>{\columncolor[HTML]{FFFFFF}}c }
\toprule
 & \textbf{Acrobot} & \textbf{CartPole} & \textbf{LunarLander} & \textbf{MountainCar} & \textbf{MountainCar Continuous} & \textbf{Pendulum-v1} \\
\midrule
NN: A2C & $-82.27 \pm 3.29$ & $500.00 \pm 0.0$ & $-43.01 \pm 106.26$ & $-148.90 \pm 24.10$ & $92.66 \pm 0.32$ & \textbf{\boldmath$-183.64 \pm 22.32$} \\
GBRL: A2C & $-90.73 \pm 2.98$ & $500.00 \pm 0.0$ & \textbf{\boldmath$47.93 \pm 41.00$} & $-124.42 \pm 5.74$ & $93.15 \pm 1.19$ & $-538.83 \pm 66.25$ \\
\midrule
NN: AWR & $-102.53 \pm 57.25$ & $500.00 \pm 0.0$ & \textbf{\boldmath$282.48 \pm 1.96$} & $-160.65 \pm 53.97$ & $18.93 \pm 42.34$ & \textbf{\boldmath$-159.64 \pm 9.42$} \\
GBRL: AWR & $-118.12 \pm 33.54$ & $497.54 \pm 3.11$ & $76.03 \pm 56.62$ & $-146.68 \pm 24.53$ & $44.38 \pm 45.94$ & $-1257.61 \pm 98.10$ \\
\midrule
NN: PPO & $-74.83 \pm 1.22$ & $500.00 \pm 0.0$ & $261.73 \pm 6.93$ & $-115.53 \pm 1.39$ & $85.81 \pm 7.51$ & $-249.31 \pm 60.00$ \\
GBRL: PPO & $-87.82 \pm 2.16$ & $500.00 \pm 0.0$ & $248.72 \pm 59.10$ & $-110.55 \pm 15.60$ & $89.42 \pm 5.73$ & $-246.89 \pm 20.61$ \\
\bottomrule
\end{tabular}%
}
\label{tab:gym}
\end{table}

\begin{table}[h!]
\centering
\caption{Football Academy environments: Reported values are the mean and standard deviation of the average episode reward over the final 100 training episodes, aggregated across 5 random seeds.}
\vspace{\baselineskip}
\begin{subfigure}{0.9\textwidth}
\centering
\resizebox{\textwidth}{!}{%
\begin{tabular}{
>{\columncolor[HTML]{FFFFFF}}c 
>{\columncolor[HTML]{FFFFFF}}c 
>{\columncolor[HTML]{FFFFFF}}c 
>{\columncolor[HTML]{FFFFFF}}c 
>{\columncolor[HTML]{FFFFFF}}c 
>{\columncolor[HTML]{FFFFFF}}c 
>{\columncolor[HTML]{FFFFFF}}c 
}
\toprule
 & \textbf{3 vs 1 with keeper} & \textbf{Corner} & \textbf{Counterattack Easy} & \textbf{Counterattack Hard} & \textbf{Empty Goal} & \textbf{Empty Goal Close} \\
\midrule
NN: A2C & $1.78 \pm 0.10$ & $1.00 \pm 0.17$ & \textbf{\boldmath$1.58 \pm 0.35$} & \textbf{\boldmath$1.43 \pm 0.17$} & \textbf{\boldmath$1.93 \pm 0.05$} & $2.0 \pm 0.0$ \\
GBRL: A2C & $1.59 \pm 0.17$ & $1.01 \pm 0.07$ & $1.11 \pm 0.14$ & $1.00 \pm 0.05$ & $1.81 \pm 0.03$ & $2.00 \pm 0.00$ \\    
\midrule
NN: AWR & $1.50 \pm 0.37$ & $1.01 \pm 0.04$ & \textbf{\boldmath$1.59 \pm 0.36$} & $1.18 \pm 0.21$ & $1.90 \pm 0.08$ & $1.92 \pm 0.17$ \\ 
GBRL: AWR & $1.66 \pm 0.34$ & $0.92 \pm 0.05$ & $0.95 \pm 0.05$ & $0.92 \pm 0.05$ & $1.93 \pm 0.07$ & $2.0 \pm 0.0$ \\
\midrule
NN: PPO & $1.61 \pm 0.05$ & $0.95 \pm 0.02$ & $1.43 \pm 0.15$ & $1.23 \pm 0.18$ & \textbf{\boldmath$1.98 \pm 0.01$} & $1.99 \pm 0.00$ \\
GBRL: PPO & $1.63 \pm 0.19$ & $1.05 \pm 0.20$ & $1.64 \pm 0.09$ & $1.23 \pm 0.07$ & $1.84 \pm 0.06$ & $2.0 \pm 0.0$ \\ 
\bottomrule
\end{tabular}%
}
\end{subfigure}

\vspace{\baselineskip}

\begin{subfigure}{0.9\textwidth}
\centering
\resizebox{\textwidth}{!}{%
\begin{tabular}{
>{\columncolor[HTML]{FFFFFF}}c 
>{\columncolor[HTML]{FFFFFF}}c 
>{\columncolor[HTML]{FFFFFF}}c 
>{\columncolor[HTML]{FFFFFF}}c 
>{\columncolor[HTML]{FFFFFF}}c 
>{\columncolor[HTML]{FFFFFF}}c 
}
\toprule
 & \textbf{Pass \& Shoot keeper} & \textbf{Run Pass \& Shoot keeper} & \textbf{Run to Score} & \textbf{Run to score w/ keeper} & \textbf{Single Goal vs Lazy} \\
\midrule
NN: A2C & $1.41 \pm 0.37$ & $1.77 \pm 0.08$ & $1.87 \pm 0.12$ & $1.25 \pm 0.23$ & \textbf{\boldmath$1.65 \pm 0.04$} \\ 
GBRL: A2C & $1.60 \pm 0.21$ & $1.60 \pm 0.14$ & $1.82 \pm 0.10$ & $1.15 \pm 0.08$ & $1.31 \pm 0.11$ \\
\midrule
NN: AWR & $1.26 \pm 0.46$ & $1.15 \pm 0.14$ & $1.81 \pm 0.14$ & $1.25 \pm 0.34$ & $1.28 \pm 0.27$ \\
GBRL: AWR & $1.35 \pm 0.37$ & $1.53 \pm 0.40$ & \textbf{\boldmath$1.98 \pm 0.01$} & $0.99 \pm 0.16$ & $1.03 \pm 0.12$ \\
\midrule
NN: PPO & $1.31 \pm 0.13$ & $1.64 \pm 0.16$ & $1.91 \pm 0.09$ & $1.13 \pm 0.06$ & $1.68 \pm 0.09$ \\
GBRL: PPO & \textbf{\boldmath$1.87 \pm 0.09$} & $1.85 \pm 0.08$ & $1.83 \pm 0.04$ & \textbf{\boldmath$1.95 \pm 0.02$} & $1.73 \pm 0.06$ \\
\bottomrule
\end{tabular}%
}
\end{subfigure}
\label{tab:football}
\vspace{\baselineskip}
\end{table}

\begin{table}[h!]
\centering
\caption{Atari-ramNoFrameskip-v4 environments: Reported values are the mean and standard deviation of the average episode reward over the final 100 training episodes, aggregated across 5 random seeds.}
\begin{subfigure}{0.9\textwidth}
\centering
\resizebox{\textwidth}{!}{%
\begin{tabular}{
>{\columncolor[HTML]{FFFFFF}}c 
>{\columncolor[HTML]{FFFFFF}}c 
>{\columncolor[HTML]{FFFFFF}}c 
>{\columncolor[HTML]{FFFFFF}}c 
>{\columncolor[HTML]{FFFFFF}}c 
>{\columncolor[HTML]{FFFFFF}}c }
\toprule
 & \textbf{Alien} & \textbf{Amidar} & \textbf{Asteroids} & \textbf{Breakout} & \textbf{Gopher} \\ 
\midrule
NN: A2C & \textbf{\boldmath$1802.24 \pm 323.12$} & \textbf{\boldmath$304.62 \pm 55.61$} & \textbf{\boldmath$2770.46 \pm 271.97$} & \textbf{\boldmath$76.69 \pm 30.08$} & \textbf{\boldmath$3533.84 \pm 118.50$} \\
GBRL: A2C & $595.08 \pm 43.51$ & $48.71 \pm 14.65$ & $1402.66 \pm 161.67$ & $11.52 \pm 2.34$ & $502.20 \pm 341.88$ \\ 
\midrule
NN: AWR & $739.82 \pm 303.06$ & $86.32 \pm 40.16$ & \textbf{\boldmath$2308.68 \pm 257.72$} & $26.57 \pm 9.91$ & $1471.93 \pm 716.65$ \\
GBRL: AWR & $829.99 \pm 166.48$ & $125.53 \pm 25.25$ & $1592.63 \pm 109.96$ & $17.32 \pm 1.89$ & $913.06 \pm 79.95$ \\
\midrule
NN: PPO & \textbf{\boldmath$1555.32 \pm 107.59$} & \textbf{\boldmath$310.93 \pm 80.13$} & \textbf{\boldmath$2309.46 \pm 145.66$} & \textbf{\boldmath$32.88 \pm 15.74$} & \textbf{\boldmath$2507.84 \pm 108.37$} \\
GBRL: PPO & $1163.86 \pm 76.54$ & $186.32 \pm 50.63$ & $1514.34 \pm 317.46$ & $19.96 \pm 1.93$ & $1215.04 \pm 81.01$ \\
\bottomrule
\end{tabular}%
}
\end{subfigure}

\vspace{\baselineskip} 

\begin{subfigure}{0.9\textwidth}
\centering
\resizebox{\textwidth}{!}{%
\begin{tabular}{
>{\columncolor[HTML]{FFFFFF}}c 
>{\columncolor[HTML]{FFFFFF}}c 
>{\columncolor[HTML]{FFFFFF}}c 
>{\columncolor[HTML]{FFFFFF}}c 
>{\columncolor[HTML]{FFFFFF}}c 
>{\columncolor[HTML]{FFFFFF}}c }
\toprule
 & \textbf{Kangaroo} & \textbf{Krull} & \textbf{MsPacman} & \textbf{Pong} & \textbf{SpaceInvaders} \\ 
\midrule
NN: A2C & \textbf{\boldmath$2137.6 \pm 425.64$} & \textbf{\boldmath$9325.38 \pm 777.12$} & \textbf{\boldmath$2007.64 \pm 116.52$} & \textbf{\boldmath$15.39 \pm 4.26$} & \textbf{\boldmath$462.30 \pm 35.56$} \\
GBRL: A2C & $948.8 \pm 483.80$ & $5291.4 \pm 433.35$ & $989.68 \pm 100.02$ & $-12.80 \pm 11.10$ & $265.36 \pm 44.64$ \\ 
\midrule
NN: AWR & $1214.8 \pm 313.42$ & $4519.78 \pm 522.11$ & $892.31 \pm 289.36$ & $-10.25 \pm 2.11$ & $842.00 \pm 130.51$ \\
GBRL: AWR & \textbf{\boldmath$1809.26 \pm 37.51$} & \textbf{\boldmath$6419.26 \pm 387.76$} & \textbf{\boldmath$1641.84 \pm 284.19$} & $-11.68 \pm 3.79$ & $397.85 \pm 566.38$ \\
\midrule
NN: PPO & $2487.4 \pm 829.65$ & $9167.3 \pm 294.30$ & $2069.22 \pm 202.48$ & $18.50 \pm 1.60$ & $479.77 \pm 65.07$ \\
GBRL: PPO & $2160.8 \pm 826.92$ & $6888.66 \pm 756.18$ & $2069.22 \pm 538.62$ & $15.40 \pm 6.55$ & $434.84 \pm 31.83$ \\
\bottomrule
\end{tabular}%
}
\end{subfigure}
\label{tab:atari}
\end{table}

\begin{table}[h!]
\centering
\caption{MiniGrid environments: Reported values are the mean and standard deviation of the average episode reward over the final 100 training episodes, aggregated across 5 random seeds.}
\begin{subfigure}{0.8\textwidth}
\centering
\resizebox{\textwidth}{!}{%
\begin{tabular}{
>{\columncolor[HTML]{FFFFFF}}c 
>{\columncolor[HTML]{FFFFFF}}c 
>{\columncolor[HTML]{FFFFFF}}c 
>{\columncolor[HTML]{FFFFFF}}c 
>{\columncolor[HTML]{FFFFFF}}c 
>{\columncolor[HTML]{FFFFFF}}c }
\toprule
 & \textbf{DoorKey-5x5} & \textbf{Empty-Random-5x5} & \textbf{Fetch-5x5-N2} & \textbf{FourRooms} & \textbf{GoToDoor-5x5} \\
\midrule
NN: A2C & $0.96 \pm 0.00$ & $0.77 \pm 0.42$ & $0.43 \pm 0.03$ & $0.62 \pm 0.19$ & $0.05 \pm 0.04$ \\
GBRL: A2C & $0.96 \pm 0.00$ & $0.96 \pm 0.00$ & $0.62 \pm 0.02$ & $0.51 \pm 0.07$ & $0.78 \pm 0.02$ \\
\midrule
NN: AWR & $0.57 \pm 0.52$ & $0.96 \pm 0.00$ & $0.90 \pm 0.26$ & $0.19 \pm 0.12$ & $0.95 \pm 0.01$ \\
GBRL: AWR & \textbf{\boldmath$0.96 \pm 0.00$} & $0.97 \pm 0.00$ & \textbf{\boldmath$0.95 \pm 0.01$} & \textbf{\boldmath$0.54 \pm 0.05$} & $0.94 \pm 0.01$ \\
\midrule
NN: PPO & $0.78 \pm 0.40$ & $0.96 \pm 0.00$ & $0.89 \pm 0.03$ & $0.53 \pm 0.03$ & $0.60 \pm 0.06$ \\
GBRL: PPO & $0.96 \pm 0.00$ & $0.96 \pm 0.00$ & \textbf{\boldmath$0.96 \pm 0.01$} & $0.56 \pm 0.04$ & \textbf{\boldmath$0.96 \pm 0.00$} \\
\bottomrule
\end{tabular}%
}
\end{subfigure}

\vspace{\baselineskip} 

\begin{subfigure}{0.7\textwidth}
\centering
\resizebox{\textwidth}{!}{%
\begin{tabular}{
>{\columncolor[HTML]{FFFFFF}}c 
>{\columncolor[HTML]{FFFFFF}}c 
>{\columncolor[HTML]{FFFFFF}}c 
>{\columncolor[HTML]{FFFFFF}}c 
>{\columncolor[HTML]{FFFFFF}}c }
\toprule
 & \textbf{KeyCorridorS3R1} & \textbf{PutNear-6x6-N2} & \textbf{RedBlueDoors-6x6} & \textbf{Unlock} \\
\midrule
NN: A2C & $0.75 \pm 0.42$ & $0.01 \pm 0.00$ & \textbf{\boldmath$0.30 \pm 0.22$} & $0.77 \pm 0.43$ \\
GBRL: A2C & $0.39 \pm 0.48$ & \textbf{\boldmath$0.18 \pm 0.018$} & $0.0 \pm 0.0$ & $0.90 \pm 0.09$ \\
\midrule
NN: AWR & $0.93 \pm 0.00$ & \textbf{\boldmath$0.60 \pm 0.13$} & $0.83 \pm 0.00$ & $0.96 \pm 0.00$ \\
GBRL: AWR & $0.94 \pm 0.00$ & $0.36 \pm 0.01$ & $0.84 \pm 0.03$ & $0.95 \pm 0.00$ \\
\midrule
NN: PPO & $0.76 \pm 0.42$ & $0.001 \pm 0.00$ & $0.17 \pm 0.40$ & $0.97 \pm 0.00$ \\
GBRL: PPO & \textbf{\boldmath$0.95 \pm 0.00$} & \textbf{\boldmath$0.44 \pm 0.19$} & \textbf{\boldmath$0.88 \pm 0.02$} & $0.97 \pm 0.00$ \\
\bottomrule
\end{tabular}%
}
\end{subfigure}

\label{tab:minigrid}
\end{table}

\section{Learning Curves}
\label{sec:appendix:training_plots}
\begin{figure*}[h]
\centering
    \includegraphics[width=0.85\linewidth]{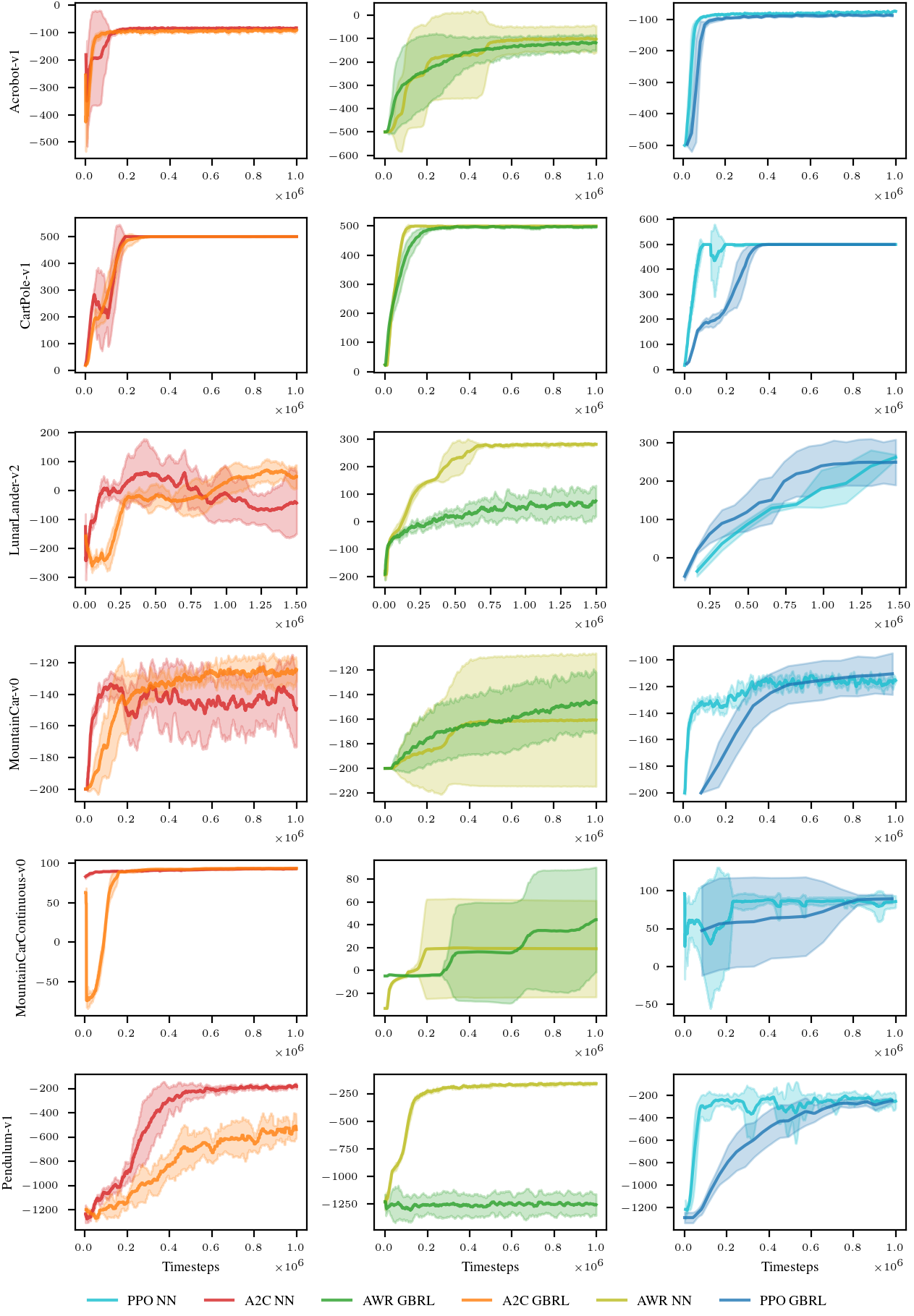}
    \caption{Classic Control and Box2D environments: Training reward as a function of environment steps. Reported values are the mean and standard deviation of the average episode reward over the final 100 training episodes, aggregated across 5 random seeds.
}
    \label{fig:gym}
\end{figure*}

\begin{figure*}[h]
\centering
    \includegraphics[width=0.85\linewidth]{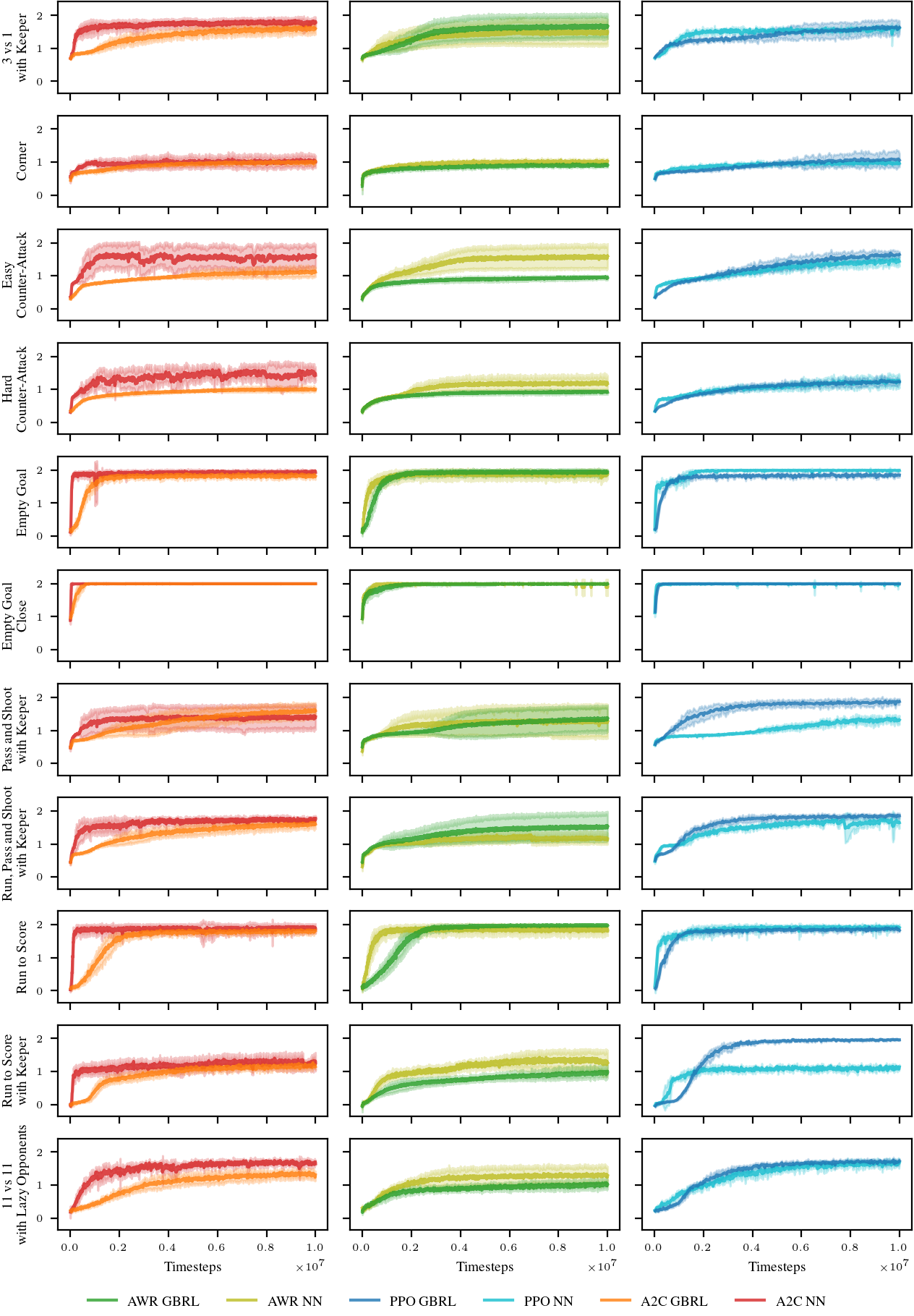}
    \caption{Football Academy environments: Training reward as a function of environment step. Reported values are the mean and standard deviation of the average episode reward over the final 100 training episodes, aggregated across 5 random seeds.
}
    \label{fig:football}
\end{figure*}

\begin{figure*}[h]
\centering
    \includegraphics[width=0.85\linewidth]{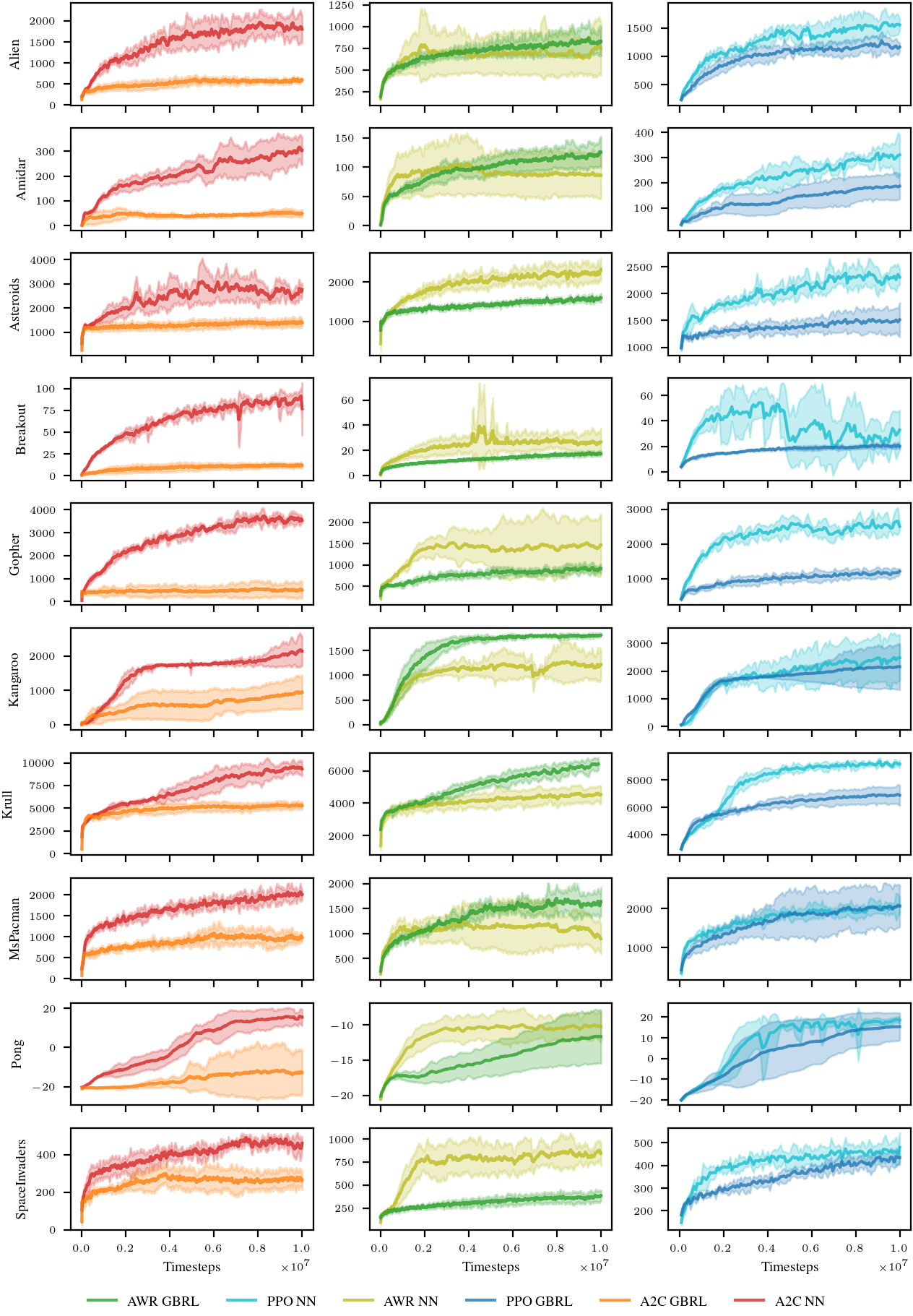}
    \caption{Atari-ramNoFrameskip-v4 environments: Training reward as a function of environment step. Reported values are the mean and standard deviation of the average episode reward over the final 100 training episodes, aggregated across 5 random seeds.
}
    \label{fig:atari}
\end{figure*}

\begin{figure*}[h]
\centering
    \includegraphics[width=0.85\linewidth]{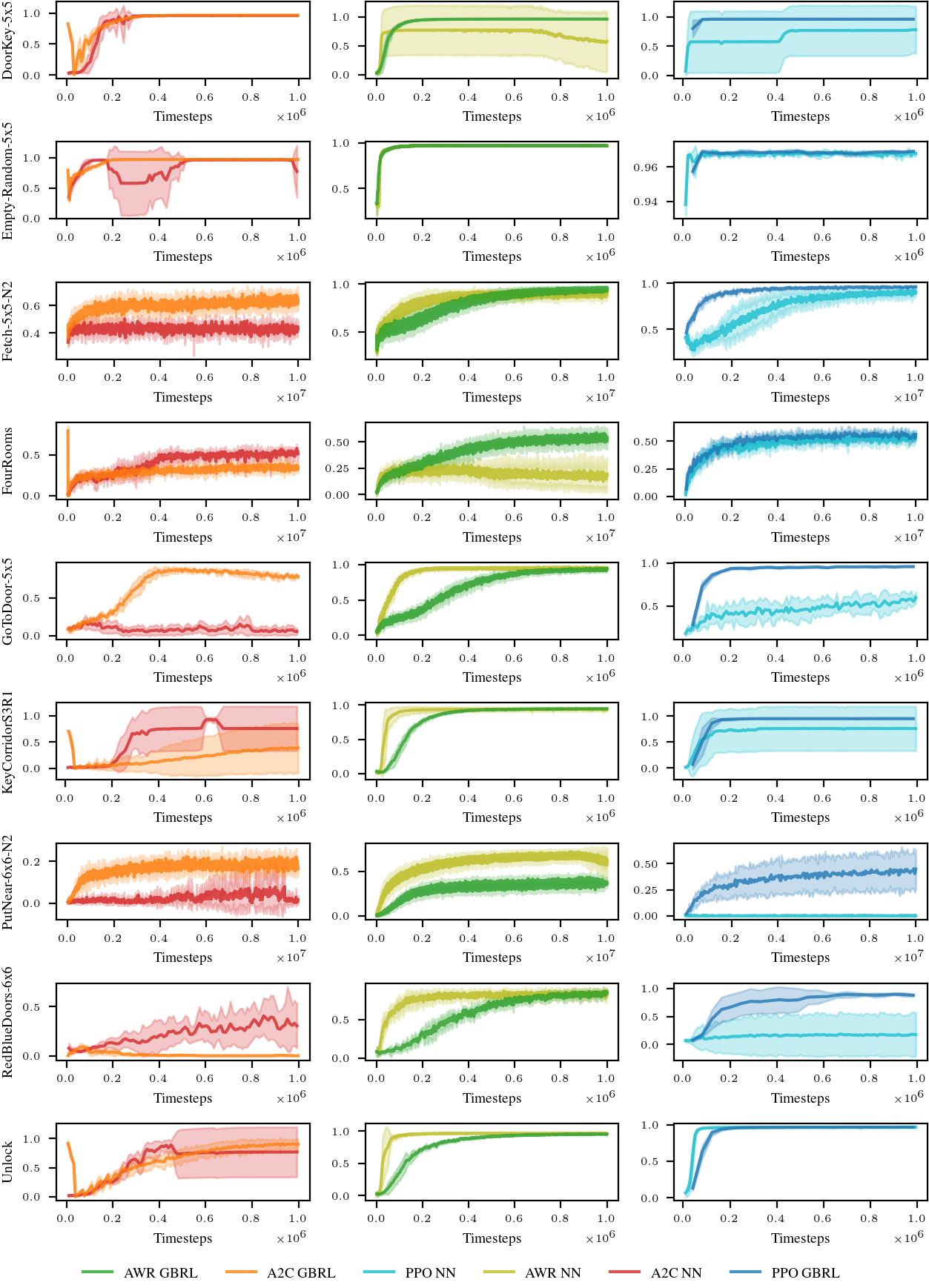}
    \caption{MiniGrid environments: Training reward as a function of environment step. Reported values are the mean and standard deviation of the average episode reward over the final 100 training episodes, aggregated across 5 random seeds.
}
    \label{fig:minigrid}
\end{figure*}

This section presents learning curves depicting model performance throughout the training phase. \cref{fig:gym,fig:football,fig:atari,fig:minigrid} show the training reward as a function of environment steps of the agents trained in the experiments. The column order is: A2C, AWR, and PPO. 

\begin{figure*}[h]
\centering
    \includegraphics[width=0.9\linewidth]{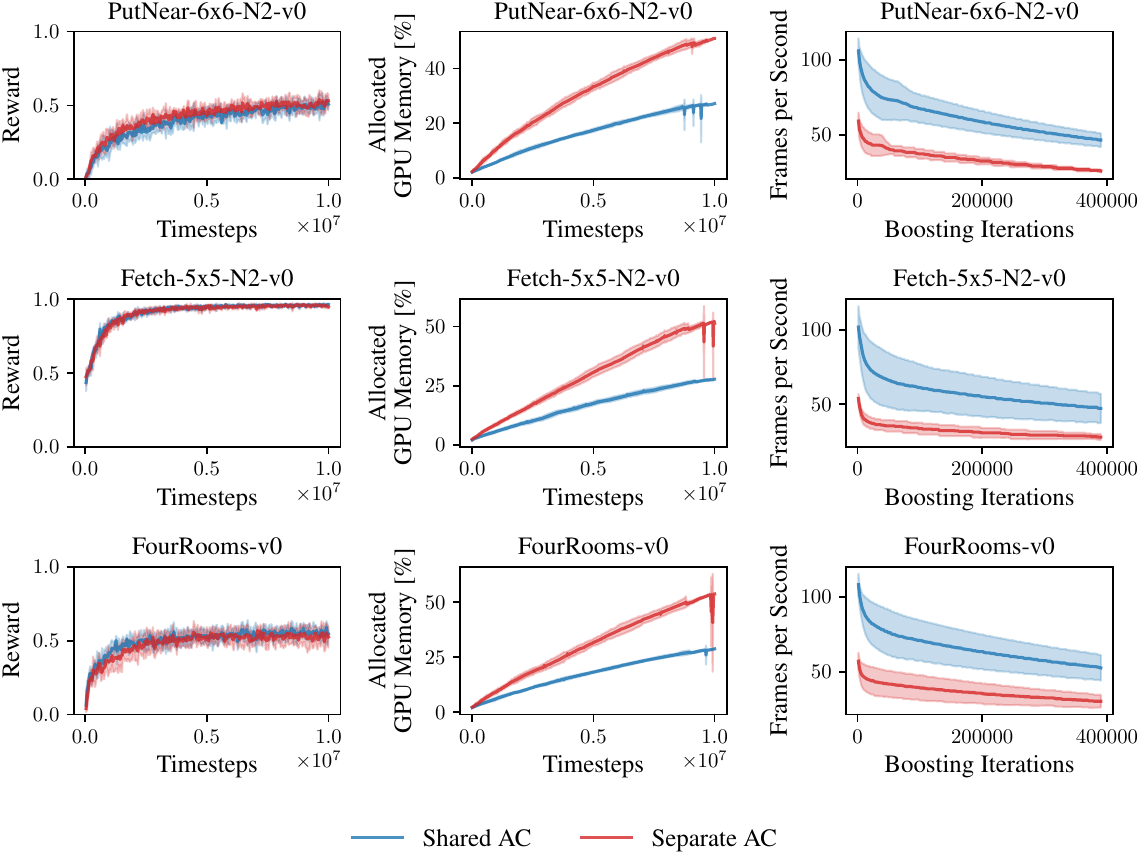}
    \caption{\textbf{Sharing actor critic tree structure significantly increases efficiency while retraining similar performance}. Training reward, GPU memory usage, and FPS, are compared across 10M environment (5 seeds, 3 MiniGrid environments)} 
    \label{fig:ablation_env}
\end{figure*}

\FloatBarrier
\newpage

\end{document}